\documentclass[11pt]{article}

\usepackage{acl}

\usepackage{times}
\usepackage{latexsym}

\usepackage[T1]{fontenc}

\usepackage[utf8]{inputenc}

\usepackage{microtype}

\usepackage{inconsolata}

\usepackage{graphicx}
\usepackage[nolist]{acronym}
\usepackage{enumitem}
\usepackage{tabularx}
\usepackage{booktabs}
\usepackage{array}
\usepackage{multirow}
\usepackage{amsmath}
\usepackage{longtable}
\usepackage{xcolor} 
\usepackage{inconsolata}
\usepackage{listings}
\usepackage[most]{tcolorbox}
\usepackage{amssymb}

\newtcblisting{mycode}[2][text]{%
  listing only,
  colback=gray!15,
  colframe=black,
  coltitle=white,
  colbacktitle=black,
  fonttitle=\bfseries,
  title={#2},
  listing options={
    language=#1,
    basicstyle=\ttfamily\footnotesize\color{black},
    breaklines=true,
    breakautoindent=true,
    breakindent=0em,
    numbers=none,
    keywordstyle=\color{black},
    commentstyle=\color{black},
    stringstyle=\color{black},
    tabsize=2
  }
}

\definecolor{lightgreen}{RGB}{90,227,58}
\definecolor{lightred}{RGB}{245,44,29}

\title{Think-Probe-Respond: Improving Large Language Models as Judges of Research Idea Novelty}

\author{
  Tim Schopf$^{1}$ \and Tobias Schreieder$^{2}$ \and Akiko Aizawa$^{1}$
  \\
  $^1$National Institute of Informatics, Tokyo, Japan \\
  $^2$TU Dresden \& ScaDS.AI Dresden/Leipzig, Germany \\
  \texttt{tim.schopf@t-online.de}, \texttt{tobias.schreieder@tu-dresden.de}, \texttt{aizawa@nii.ac.jp}
}

\begin{acronym}
    \acro{llm}[LLM]{large language model}
    \acro{mae}[MAE]{Mean Absolute Error}
    \acro{rino}[\mbox{\texttt{RINoBench}}]{\textbf{R}esearch \textbf{I}dea \textbf{No}velty Judgment \textbf{Bench}mark}
    \acro{sota}[SOTA]{state-of-the-art}
    \acro{tpr}[TPR]{\textbf{T}hink-\textbf{P}robe-\textbf{R}espond}
    \acro{cot}[CoT]{Chain-of-Thought}
\end{acronym}

\begin{document}
\maketitle
\begin{abstract}
Automated novelty judgment can accelerate scientific discovery by enabling efficient evaluation, refinement, and comparison of research ideas. While large language models are increasingly adopted for this task, we investigate a previously overlooked limitation in their judgment capabilities: despite generating reasoning rationales that closely mirror those of human experts, their final novelty judgments often diverge substantially. We demonstrate that this miscalibration stems from a systematic bias towards judging ideas as ``medium novel''. 
To mitigate this, we propose \textbf{T}hink-\textbf{P}robe-\textbf{R}espond (TPR), a lightweight approach that probes latent novelty judgments from hidden states during the reasoning phase and uses the probed judgments to condition the final response. Across strong baselines, TPR improves novelty judgment performance by 22.30\% and successfully mitigates the prevalent ``medium novelty'' bias.
\end{abstract}

\begin{figure*}[ht!]
    \centering
    \includegraphics[width=1\textwidth]{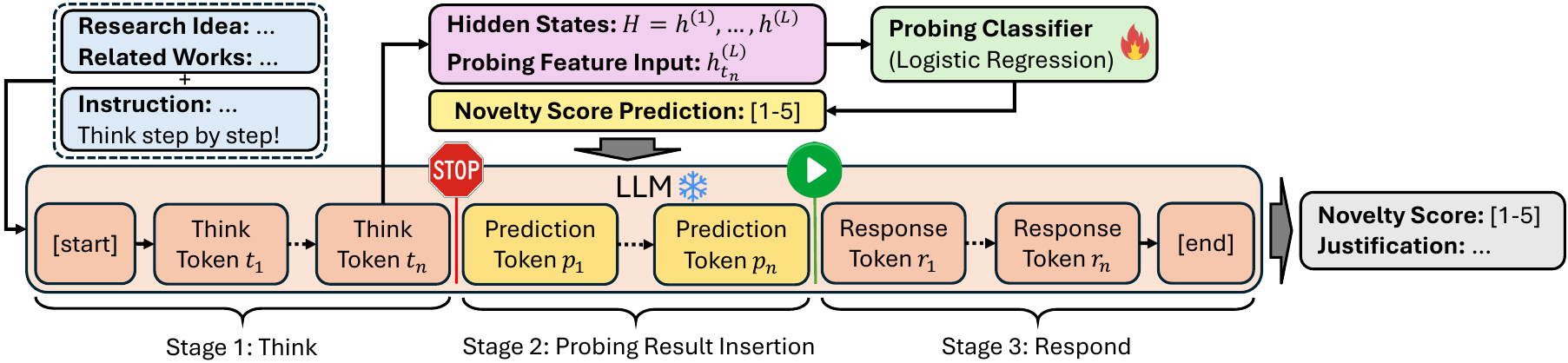}
    \caption{Overview of our \acl{tpr} approach. The snowflake (\mbox{\raisebox{-0.25em}{\includegraphics[height=1em]{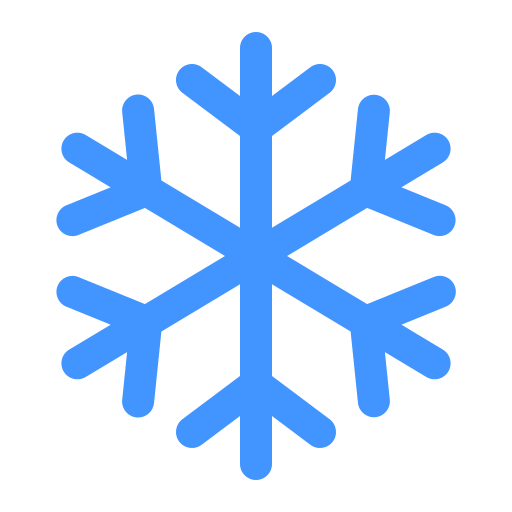}}}) indicates frozen parameters, while the flame (\mbox{\raisebox{-0.2em}{\includegraphics[height=1em]{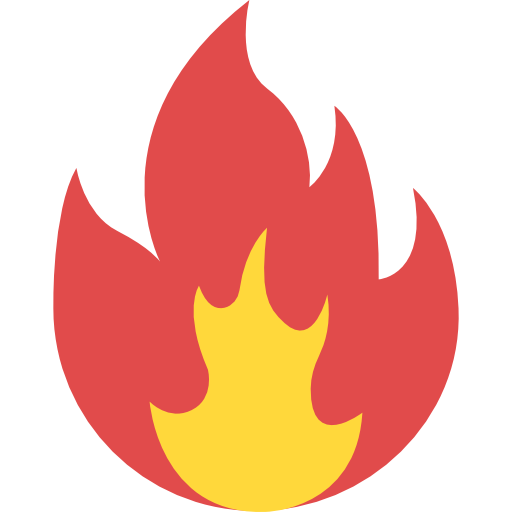}}}) indicates trainable parameters.}
    \label{fig:TPR}
\end{figure*}

\section{Introduction}

Judging the novelty of research ideas is crucial to scientific progress, enabling original contributions and shaping future scientific directions. However, manual novelty judgment requires substantial expertise and a broad understanding of the relevant literature, making it time-consuming, subjective, and difficult to scale \cite{picard25}. As scientific output continues to grow rapidly \cite{doi:10.1126/science.aao0185}, automated support for novelty judgment becomes important for helping researchers assess, refine, and compare research ideas.

Recent work increasingly relies on \acp{llm} to judge research idea novelty \cite[\textit{inter alia}]{lu2024aiscientistfullyautomated,li2024chainideasrevolutionizingresearch,ICLR2025_ea94957d,su-etal-2025-many,Lu2026,Gottweis2026,mostafa2026novelknowledgedrivenframeworkbiasaware,wu2026novbenchevaluatinglargelanguage}. 
However, a critical discrepancy persists: while these models produce plausible, human-like novelty arguments \cite{afzal-etal-2026-beyond}, their final novelty judgments remain fundamentally disconnected from their own reasoning rationales and fail to align with human judgments \cite{ICLR2025_ea94957d,schopf-etal-2026-is}.

In this work, we investigate the underlying causes of this miscalibration and demonstrate that \acp{llm} exhibit a systematic bias toward conservative ``medium novelty'' judgments, even when their internal reasoning supports substantially different judgments. Motivated by this finding, we propose \ac{tpr}, a lightweight method for mitigating novelty miscalibration. \ac{tpr} probes latent novelty beliefs from hidden states during reasoning and conditions the final response on these extracted beliefs. Experimental results show that \ac{tpr} improves novelty judgment performance by 22.30\% on average over strong baselines while producing less biased novelty judgments.

\section{Related Work}

Early efforts for automated research idea novelty judgment have evolved from citation- and lexical-based methods \cite{doi:10.1126/science.1240474,WANG20171416,amplayo-etal-2019-evaluating,wang2019idea,SARICA2020112995} to semantic embedding approaches \cite{gómezpérez2022artificialintelligencenaturallanguage}. While these methods improve semantic matching, they largely remain limited to surface-level similarity estimation \cite{mysore-etal-2022-multi}. More recent work adopt \acp{llm} for automated novelty judgments \cite[\textit{inter alia}]{https://doi.org/10.1002/asi.70005,ICLR2025_ea94957d,liu2025harnessinglargelanguagemodels,wang2025scipipllmbasedscientificpaper,baek-etal-2025-researchagent,lin-etal-2025-evaluating,zhang2025noveltybench,tang2025airesearcher,li-etal-2025-chain-ideas,feng2025grapheval,hou2026noveltyagentautonomousnoveltyreporting}. 
In contrast to prior work, we investigate a previously overlooked failure mode of \acp{llm}: the systematic miscalibration between their generated novelty rationales and final novelty judgments.

\section{Benchmark}\label{sec:benchmark}

We conduct all experiments on \acs{rino} \cite{schopf-etal-2026-is}, the only publicly available benchmark for research idea novelty judgment with human-annotated novelty scores. It comprises 1,381 expert-authored ideas, each paired with related works, a human-annotated novelty score on a five-point Likert scale (rubric in Table~\ref{tab:novelty-rubric}), and an expert-written justification supporting the assigned novelty score. Given a research idea and its related work, models must predict the novelty score ranging from 1 (not novel) to 5 (highly novel) and generate a textual justification grounded in comparisons with prior work. 

\begin{table*}[t!]
\centering
\footnotesize
\renewcommand{\arraystretch}{0.85}
\resizebox{\textwidth}{!}{
\begin{tabular}{lccccccc|ccccccc}
\toprule
 & \multicolumn{6}{c}{$F_1$} & MAE & ALI & \multicolumn{2}{c}{Recall} & \multicolumn{2}{c}{Add. Ratio} & \multicolumn{2}{c}{Hall. Rate} \\
\cmidrule(lr){2-7} \cmidrule(lr){10-11} \cmidrule(lr){12-13} \cmidrule(lr){14-15}
Model & Macro & 1 & 2 & 3 & 4 & 5 & & & KA & NA & KA & NA & KA & NA \\
\midrule
Gemini 2.5 Pro & 15.6 & 0.0 & 6.3 & 28.8 & \textbf{42.9} & 0.0 & 1.0 & 0.41 & 64.7 & 60.6 & 104.6 & 96.5 & 8.2 & 1.2 \\
Gemini 3 Pro & 15.3 & \textbf{9.8} & \textbf{30.9} & 35.8 & 0.0 & \textbf{9.8} & 1.0 & 0.46 & 64.9 & 67.0 & 101.9 & 81.2 & 13.8 & 3.2 \\
Claude Sonnet 4.5 & 15.5 & 0.0 & 14.6 & 43.2 & 19.5 & 0.0 & 0.9 & 0.59 & 80.0 & \textbf{70.3} & \textbf{150.1} & 14.8 & 8.7 & 1.2 \\
Claude Opus 4.5 & \textbf{17.1} & 0.0 & 12.1 & 41.6 & 31.9 & 0.0 & \textbf{0.8} & 0.61 & \textbf{80.4} & 68.8 & 143.4 & \textbf{106.0} & 8.0 & 1.2 \\
GPT-5 mini & 16.2 & 0.0 & 5.1 & 40.8 & 34.9 & 0.0 & 0.9 & \textbf{0.62} & 77.3 & 67.8 & 144.8 & 104.8 & 7.8 & 1.0 \\
GPT-5.4 & 15.1 & 0.0 & 3.0 & \textbf{48.3} & 24.3 & 0.0 & \textbf{0.8} & \textbf{0.62} & 75.8 & 67.8 & 117.2 & 77.3 & \textbf{4.8} & \textbf{0.6} \\
\bottomrule
\end{tabular}
}
\caption{Evaluation results of novelty judgments on the \acs{rino} test set. The reported metrics include \textit{$F_{1}$} macro averaged and for each rubric category (1-5), \textit{\ac{mae}}, \textit{Alignment (ALI)}, \textit{Recall}, \textit{Additional Ratio} (in \%), and \textit{Hallucination Rate} (in \%) for \textit{Known Aspects (KA)} and  \textit{Novelty Aspects (NA)} respectively (for more details on the metrics, see Appendix \ref{sec:eval-metrics}).}
\label{tab:benchmarking_results}
\end{table*}

\section{Miscalibration in LLM Novelty Judgments}\label{sec:benchmarking}

Although \ac{llm} novelty rationales align closely with human reasoning \cite{afzal-etal-2026-beyond}, their final novelty judgments often diverge significantly from human consensus \cite{ICLR2025_ea94957d,schopf-etal-2026-is}. To investigate this miscalibration and examine \textit{why} \acp{llm} struggle to produce accurate novelty judgments, we prompt six \ac{sota} \acp{llm} (Gemini~2.5~Pro \cite{comanici2025gemini25pushingfrontier}, Gemini~3~Pro \cite{gemini-3}, Claude~Sonnet~4.5 \cite{anthropicIntroducingClaudeSonnet}, Claude~Opus~4.5 \cite{anthropicIntroducingClaudeOpus}, GPT-5 mini \cite{singh2026openaigpt5card}, GPT-5.4 \cite{OpenAI-gpt-5.4}) to judge research idea novelty (prompt in Figure~\ref{fig:zs-prompt}). As Table~\ref{tab:benchmarking_results} shows, all models perform poorly on the novelty judgment task, yielding \ac{mae} values around one and consistently low macro-\textit{$F_{1}$} scores, with the best-performing model achieving just 17.1.

\noindent\textbf{LLMs Avoid Extreme Novelty Judgments} The dominant failure pattern is a strong middle-ground bias. Across models, predictions concentrate on novelty classes 3 and 4, while the lowest and highest categories are rarely predicted correctly. With the exception of Gemini~3~Pro, which occasionally assigns classes~1 and~5, models fail almost entirely on these extreme cases. Table \ref{tab:novelty_comparison} shows examples of this \ac{llm} behavior. 

\noindent\textbf{Latent Belief vs.\ Expressed Novelty Judgment} This middle-ground bias contrasts with the quality of the generated justifications. Models achieve high recall with respect to human-annotated justification arguments, indicating that they often identify the same overlaps, differences, and novelty aspects as human experts. 
Overall, these results suggest that \acp{llm} often \textit{``know''} substantially more about the true novelty level than their generated numerical novelty scores reveal.


\noindent\textbf{Takeaway} The miscalibration of \acp{llm} as novelty judges is therefore best understood as a mismatch between latent belief and expressed judgment. \textit{The models' reasoning often contains evidence for accurate novelty judgment, but their final outputs are biased towards safe middle categories.} 

\section{Probing Latent LLM Judgments}\label{sec:tpr}

Building on the finding that \acp{llm} internalize beliefs about research idea novelty that closely mirror those of human experts—and therefore generate comparable novelty rationales—yet are biased towards predicting medium novelty categories, we propose the \ac{tpr} approach for research idea novelty judgment. \ac{tpr} explicitly exploits the model’s internal beliefs during reasoning about novelty to yield less biased and more accurate quantitative judgments. These judgments are then reused as conditioning signals to generate textual justifications that are coherent and well-aligned with the predicted numerical scores. As illustrated in Figure \ref{fig:TPR}, \ac{tpr} consists of three stages. (1) \textbf{Think:} We instruct an \ac{llm} to judge the novelty of a research idea and to think step by step before producing a final response. For reasoning models that generate think tokens by default, we omit any explicit “think step by step” instruction. Importantly, we provide only textual descriptions of the novelty categories—without numerical scores—and instruct the model to evaluate novelty solely based on these descriptions without generating numerical judgments. This design encourages the model to think about the novelty of research ideas qualitatively, avoiding anchoring its internal representations to explicit numeric outcomes that could bias the reasoning process. 
Figure \ref{fig:tpr-prompt} shows the prompt used in this approach. (2) \textbf{Probe:} Given an LLM with $L$ hidden layers, let $H = h^{(1)}, \dots, h^{(L)}$ represent the stack of hidden states. For a generated sequence of reasoning (“think”) tokens $T = t_{1}, \dots, t_{n}$, we terminate generation upon the production of the final think token $t_{n}$ and extract the hidden state $h^{(L)}_{t_{n}}$ (Section \ref{sec:probing-tokens} motivates the choice of $t_{n}$). This representation $h^{(L)}_{t_{n}}$ is then used as the input feature vector for a logistic regression probing classifier.\footnote{While probing intermediate layers is a viable alternative, we follow prior work showing that representations from the final layer typically yield strong probing performance and offer practical advantages, as extraction of the last hidden state of a given model is more easily facilitated in common open-\ac{llm} frameworks than earlier layers \cite{maiya-etal-2025-improving}.} (3) \textbf{Respond:} We append the textual description of the predicted novelty class to the \ac{llm}-generated output and resume generation. Conditioned on both its prior reasoning and the predicted novelty judgment, the \ac{llm} generates the final response, which is used as the justification of the novelty judgment. 

\ac{tpr} is computationally efficient and lightweight. All \ac{llm} parameters remain frozen and are used only at inference. Training is limited to a simple logistic regression classifier, which can be learned efficiently on CPU.

\subsection{Experimental Setup}\label{sec:tpr-experiments}

We evaluate \ac{tpr} against a diverse set of approaches. These include \textbf{Zero-shot} prompting, as used in Section~\ref{sec:benchmarking}; \textbf{Few-shot} prompting adapted from \cite{shahid-etal-2025-literature}, which provides one example per novelty class; \textbf{\ac{cot}} prompting, where the model is instructed to reason step by step before producing a novelty judgment; several prompt-based methods derived from \textbf{Moose} \cite{yang-etal-2024-large-language}, \textbf{ResearchAgent} \cite{baek-etal-2025-researchagent}, \textbf{AI Scientist} \cite{lu2024aiscientistfullyautomated}, and \textbf{AI Researcher} \cite{ICLR2025_ea94957d}; as well as a \textbf{FineTune} approach that uses LoRA \cite{hu2022lora} for \ac{llm} fine-tuning  (training details are provided in Appendix \ref{sec:fine-tune}).

Since we require access to model parameters, we conduct experiments exclusively with open-source \acp{llm} spanning multiple model families. We evaluate reasoning models that generate explicit think tokens by default, namely Qwen3 (4B, 14B, 32B) \cite{yang2025qwen3technicalreport} and GPT-OSS-20B \cite{openai2025gptoss120bgptoss20bmodel}, as well as non-reasoning models for which we explicitly instruct step-by-step reasoning to elicit think tokens under \ac{tpr}, including Gemma~3 (4B, 12B, 27B) \cite{gemmateam2025gemma3technicalreport} and Llama~3.1 (8B, 70B) \cite{grattafiori2024llama3herdmodels}.

\subsection{Evaluation Results}\label{sec:tpr-evaluation}

\begin{figure*}
    \centering
    \includegraphics[width=1\linewidth]{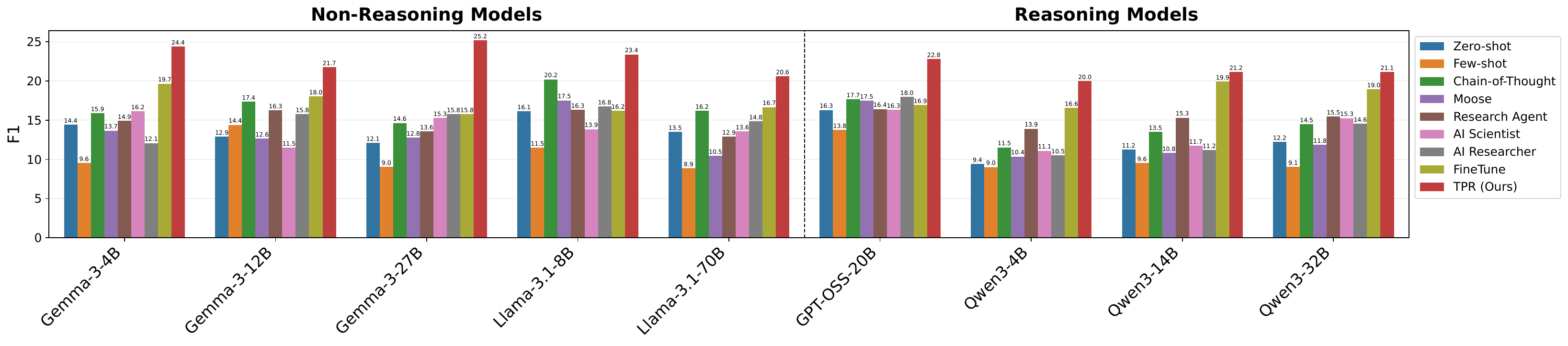}
    \caption{Macro-$F_{1}$ scores for different approaches and \acp{llm} on the \acs{rino} test set.}
    \label{fig:f1_macro_scores}
\end{figure*}

Figure~\ref{fig:f1_macro_scores} presents macro-$F_1$ scores for different \acp{llm} and approaches (see Appendix~\ref{sec:metrics-choice} for metric selection details). Across all evaluated models, \ac{tpr} achieves the highest performance, surpassing the best competing approach by an average of 22.30\%. 
Remarkably, this performance is achieved using only a lightweight logistic regression probing classifier, significantly surpassing the computationally expensive FineTune approach. While FineTune often outperforms prompt-based approaches, it cannot match the performance of \ac{tpr}.

\noindent\textbf{Consistency Across Prompt-Based Approaches} The effectiveness of prompt-based methods varies substantially across models, with no single prompting strategy consistently outperforming others. In contrast, \ac{tpr} demonstrates robust, model-agnostic performance, delivering strong results across all \acp{llm} investigated

\noindent\textbf{Model Size Is Not Determinative} Probing \ac{llm} beliefs during reasoning yields strong novelty judgments regardless of model size. Larger models do not necessarily perform better: for instance, the biggest Llama-3.1-70B ranks second-worst, whereas the smallest Gemma3-4B ranks second-best. Moreover, all open-source models using \ac{tpr} surpass the novelty judgment macro-$F_{1}$ scores of proprietary models in Table~\ref{tab:benchmarking_results}. This indicates that even smaller models using \ac{tpr} can outperform prompting substantially larger \acp{llm}.

\noindent\textbf{Reasoning vs.\ Non-Reasoning Models} \ac{tpr} is effective for both reasoning and non-reasoning models, showing that think tokens encode useful information about novelty judgments, whether generated automatically or via explicit instruction. Interestingly, reasoning models exhibit slightly lower \ac{tpr} performance on average than non-reasoning models. Fine-tuning reasoning models such as Qwen3 provides modest gains, yet \ac{tpr} still outperforms FineTune, albeit with a smaller margin than for non-reasoning models. Overall, the largest gains appear when \ac{tpr} is applied to non-reasoning models with explicit "think step-by-step" instructions.

\noindent\textbf{\ac{tpr} vs.\ \ac{cot}} Across all \acp{llm}, \ac{tpr} substantially outperforms the \ac{cot} approach. While \ac{cot} marginally benefits non-reasoning models, its effect on reasoning models is inconsistent, and its overall performance lags behind both FineTune and \ac{tpr}. These results suggest that merely instructing \acp{llm} to reason is insufficient; explicitly probing the hidden representations formed during the reasoning process is crucial for achieving high-quality novelty judgments.

\noindent\textbf{\ac{tpr} Enables Balanced Predictions Across Novelty Classes} Table~\ref{tab:class-wise-f1-tpr} reports class-wise $F_1$ scores for different \acp{llm} using our \ac{tpr} approach, while Table~\ref{tab:benchmarking_results} shows the corresponding scores for proprietary LLMs under zero-shot prompting. Compared to the prompting approach, \ac{tpr} produces substantially more balanced predictions across all novelty classes. In particular, prompted proprietary \acp{llm} often avoid extreme novelty judgments (classes 1 and 5), whereas \ac{tpr} enables models to recognize both very low and very high research idea novelties. The exception is the Qwen3 model family, which rarely assigns class 1 even under \ac{tpr}, indicating a persistent tendency to avoid “no novelty” judgments. Overall, these results show that \ac{tpr} not only improves macro-level $F_1$ performance but also encourages more uniform coverage of the full novelty spectrum, mitigating the middle-class novelty judgment bias observed in Section \ref{sec:benchmarking}.

\begin{table}[h!]
\centering
\footnotesize
\renewcommand{\arraystretch}{0.9}
\resizebox{\columnwidth}{!}{
\begin{tabular}{llcccccc}
\toprule
 & & \multicolumn{5}{c}{$F_1$}  \\
\cmidrule(lr){3-8}
& Model & Macro & 1 & 2 & 3 & 4 & 5 \\
\midrule
\multirow{5}{*}{\rotatebox{90}{Non-Reas.}}
& Gemma-3-4B & 24.4 & \textbf{27.3} & 21.5 & 35.11 & 29.71 & 8.3 \\
& Gemma-3-12B & 21.8 & 7.7 & 16.3 & 34.1 & 35.0 & 15.7 \\
& Gemma-3-27B & \textbf{25.2} & 18.2 & 19.8 & 28.9 & 33.5 & \textbf{25.4} \\
& Llama-3.1-8B & 23.4 & 14.3 & 26.0 & 36.0 & 31.2 & 9.5 \\
& Llama-3.1-70B & 20.6 & 7.7 & \textbf{27.9} & 33.9 & 24.1 & 9.5 \\
\midrule
\multirow{4}{*}{\rotatebox{90}{Reas.}} 
& Qwen3-4B & 20.0 & 0.0 & 24.4 & 31.2 & 28.2 & 16.1 \\
& Qwen3-14B & 21.2 & 0.0 & 27.1 & 33.5 & \textbf{37.4} & 7.8 \\
& Qwen3-32B & 21.1 & 0.0 & 23.5 & \textbf{36.5} & 35.8 & 9.8 \\
& GPT-OSS-20B & 22.8 & 10.0 & 24.1 & 35.7 & 34.1 & 10.0 \\
\bottomrule
\end{tabular}
}
\caption{$F_{1}$ scores per novelty class using \ac{tpr} with different \acp{llm}.}
\label{tab:class-wise-f1-tpr}
\end{table}

An additional evaluation of the textual justifications is provided in Appendix \ref{sec:justification-evaluation}.


\section{Probing over Time}\label{sec:probing-tokens}

\begin{table*}[ht!]
\centering
\renewcommand{\arraystretch}{0.85}
\resizebox{\textwidth}{!}{
\begin{tabular}{llcccccc|cccccc}
\toprule
 & & \multicolumn{6}{c}{Thinking Phase} & \multicolumn{6}{c}{Response Generation Phase} \\
\cmidrule(lr){3-8} \cmidrule(lr){9-14}
 & Model & $t_{1}$ & $t_{25\%}$ & $t_{50\%}$ & $t_{75\%}$ & $t_{n}$  & Avg. & $r_{1}$ & $r_{25\%}$ & $r_{50\%}$ & $r_{75\%}$ & $r_{n}$ & Avg. \\
\midrule
\multirow{5}{*}{\rotatebox{90}{Non-Reason.}} 
 & Gemma-3-4B & 22.74 & 16.26 & 15.47 & 16.86 & \textbf{24.38} & 19.14 & 20.84 & 18.09 & 19.86 & 17.94 & 19.81 & \underline{19.31} \\
 & Gemma-3-12B & 22.34 & 22.12 & 17.60 & 21.12 & 21.74 & \underline{20.98} & 17.87 & 16.23 & \textbf{23.73} & 20.53 & 17.20 & 19.11 \\
 & Gemma-3-27B & 18.88 & 21.36 & 18.42 & 21.51 & \textbf{25.17} & \underline{21.07} & 21.90 & 16.98 & 15.48 & 19.83 & 14.96 & 17.83 \\
 & Llama-3.1-8B & 17.01 & 19.90 & 19.77 & 19.94 & \textbf{23.38} & \underline{20.00} & 17.92 & 19.47 & 17.40 & 17.37 & 19.51 & 18.33 \\
 & Llama-3.1-70B & 21.11 & 19.12 & 20.16 & 15.42 & 20.61 & \underline{19.28} & 15.82 & 21.56 & 18.49 & \textbf{20.78} & 16.50 & 18.63 \\
\midrule
\multirow{4}{*}{\rotatebox{90}{Reason.}} 
 & Qwen3-4B & 19.06 & 19.23 & 20.51 & \textbf{21.37} & 19.98 & \underline{20.03} & 16.89 & 19.25 & 17.84 & 18.89 & 17.78 & 18.13 \\
 & Qwen3-14B & 19.05 & 16.76 & 19.65 & 16.76 & \textbf{21.17} & 18.68 & 20.47 & 18.56 & 19.77 & 19.81 & 19.49 & \underline{19.62} \\
 & Qwen3-32B & 17.51 & 17.77 & 19.76 & 22.26 & 21.13 & 19.69 & 19.32 & 19.30 & 20.20 & \textbf{22.01} & 20.07 & \underline{20.18} \\
 & GPT-OSS-20B & 22.65 & 17.02 & 19.00 & 15.85 & \textbf{22.78} & 19.46 & 19.68 & 20.00 & 22.24 & 19.21 & 18.43 & \underline{19.91} \\
\bottomrule
\end{tabular}
}
\caption{Macro $F_{1}$ scores for research idea novelty judgments on the \acs{rino} test set obtained by probing last-layer hidden states of various \acp{llm} at different generation steps: when generating the first and last think tokens ($t_{1}$, $t_{n}$), the first and last response tokens ($r_{1}$, $r_{n}$), and intermediate steps during both the thinking and response generation phases (reported as percentages of token generation within each phase; e.g., $t_{50\%}$ denotes probing halfway through the thinking phase, after 50\% of think tokens have been generated).}
\label{tab:probing_time}
\end{table*}

We investigate at which generation stages \acp{llm} encode the most salient information for research idea novelty judgments when applying \ac{tpr}. For this analysis, we allow the models to generate their full reasoning and response exactly as instructed by the prompt in Figure \ref{fig:tpr-prompt}, \textit{without} inserting the probing classifier’s prediction as an intermediate control signal. This setup enables a clean examination of when novelty-related information naturally emerges in the model’s representations. We probe last-layer hidden states at multiple time steps during both the thinking and response generation phases and report the results in Table~\ref{tab:probing_time}.

\noindent\textbf{Novelty Signals Peak at the End of the Thinking Phase} Across nearly all models, probing at the \textit{end of the thinking phase} ($t_n$) yields the strongest or near-strongest novelty judgment performance. This pattern holds consistently for both reasoning and non-reasoning models, indicating that novelty-related beliefs are most fully consolidated once the model has completed its internal reasoning process. In contrast, probing earlier thinking tokens (e.g., $t_{25\%}$ or $t_{50\%}$) generally results in substantially lower performance, suggesting that novelty representations emerge progressively rather than being present at the onset of reasoning.

\noindent\textbf{Probing Is Robust Across Reasoning Lengths} As shown in Table~\ref{tab:token_numbers}, models vary widely in the number of tokens generated during the thinking phase, from short sequences to long reasoning chains. Despite these differences, probing at $t_n$ provides consistently strong novelty signals. Importantly, there is no clear correlation between the absolute length of reasoning chains and probing performance: models with shorter or longer chains achieve comparable $F_1$ scores at the final thinking token. This indicates that the position within the reasoning sequence (final token) matters more than absolute length for capturing novelty-related beliefs.

\begin{table}[ht!]
\centering
\renewcommand{\arraystretch}{0.85}
\resizebox{\columnwidth}{!}{
\begin{tabular}{llccc|ccc}
\toprule
 & & \multicolumn{3}{c}{\# Think Tokens} & \multicolumn{3}{c}{\# Response Tokens} \\
\cmidrule(lr){3-5} \cmidrule(lr){6-8}
 & Model & Min & Max & Avg. & Min & Max & Avg. \\
\midrule
\multirow{5}{*}{\rotatebox{90}{Non-Reason.}} 
 & Gemma-3-4B & 380 & 1572 & 741.98 & 21 & 128 & 80.69 \\
 & Gemma-3-12B & 233 & 1672 & 631.21 & 44 & 105 & 69.25 \\
 & Gemma-3-27B & 206 & 804 & 431.64 & 48 & 107 & 71.81  \\
 & Llama-3.1-8B & 17 & 4358 & 435.33 & 21 & 152 & 67.79 \\
 & Llama-3.1-70B & 18 & 1034 & 313.76 & 23 & 120 & 57.23 \\
\midrule
\multirow{4}{*}{\rotatebox{90}{Reason.}} 
 & Qwen3-4B & 312 & 2525 & 1055.51 & 45 & 193 & 125.74  \\
 & Qwen3-14B & 256 & 3076 & 674.28 & 53 & 142 & 92.48 \\
 & Qwen3-32B & 275 & 2127 &  633.75 & 65 & 155 & 106.49 \\
 & GPT-OSS-20B & 7 & 265 & 57.66 & 21 & 231 & 126.13 \\
\bottomrule
\end{tabular}
}
\caption{Number of tokens generated by different \acp{llm} during research idea novelty prediction using our \ac{tpr} approach. Note, we use GPT-OSS-20B with reasoning level \textit{``low''}, resulting in a small number of think tokens.}
\label{tab:token_numbers}
\end{table}

\noindent\textbf{Response Generation Dilutes Novelty Representations} While some models achieve local maxima at intermediate response steps (e.g., $r_{50\%}$ or $r_{75\%}$), probing during the response generation phase produces competitive but generally weaker results than probing at $t_{n}$. 
This suggests that once the model transitions to response generation, the representations become increasingly influenced by surface realization and linguistic planning, diluting the underlying novelty signal. This trend holds for both reasoning and non-reasoning models. While some models exhibit local peaks at intermediate response steps, the final thinking token remains the most reliable and stable probing point overall.

\noindent\textbf{Takeaway} \textit{Novelty judgments are primarily formed during the reasoning phase and are most reliably captured toward their later stages}.

\section{Conclusion}

We showed that \acp{llm} are miscalibrated judges of research idea novelty. Although their rationales often align with human reasoning, their final judgments are biased towards medium novelty. To mitigate this, we proposed \ac{tpr}, a lightweight approach that probes latent novelty judgments from hidden states during reasoning and conditions the final response on the probed judgment. Experiments demonstrate that \ac{tpr} improves novelty judgment performance over strong baselines by 22.30\% and reduces the prevalent medium-novelty bias.


\section{Limitations}

Our experiments are conducted on \acs{rino}, which focuses on machine learning research ideas and may not fully reflect novelty judgments in other scientific domains. In addition, \ac{tpr} requires access to model hidden states, limiting its direct applicability to closed-source \acp{llm}. Finally, novelty judgments are inherently subjective, and even expert annotations may reflect individual preferences or incomplete knowledge of the literature.

\section{Ethical Considerations}
We emphasize that this work is intended for research and educational purposes only. Users should not use our models or approaches to make formal or high-stakes judgments of research ideas, as novelty judgments are inherently subjective and context-dependent.

Our work is intended to advance AI-assisted scientific discovery by enabling models to reason about and explain novel contributions in research. However, automated predictions of research idea novelty should not replace human expert judgment. Our approaches are intended as tools to support, rather than replace human judgments of research ideas.

\section*{Acknowledgments}
This work is supported by a scholarship of the German Academic Exchange Service (DAAD) - 57557629 and by the BMFTR through a Software Campus project with identification number 16|S23070.

The authors acknowledge the financial support by the Federal Ministry of Research, Technology and Space of Germany (BMFTR) and by Sächsische Staatsministerium für Wissenschaft, Kultur und Tourismus in the programme Center of Excellence for AI-research „Center for Scalable Data Analytics and Artificial Intelligence Dresden/Leipzig“, project identification number: ScaDS.AI.

We used AI-based assistance tools to support language editing, minor formatting, and coding tasks. These tools did not contribute to the intellectual content or scientific conclusions. All content was reviewed by the authors, who assume full responsibility for the publication.


\bibliography{custom}

@article{picard25,
	author = {Picard, Cyril and Edwards, Kristen M. and Doris, Anna C. and Man, Brandon and Giannone, Giorgio and Alam, Md Ferdous and Ahmed, Faez},
	date = {2025/07/01},
	doi = {10.1007/s10462-025-11290-y},
	id = {Picard2025},
	isbn = {1573-7462},
	journal = {Artificial Intelligence Review},
	number = {9},
	pages = {288},
	title = {From concept to manufacturing: evaluating vision-language models for engineering design},
	url = {https://doi.org/10.1007/s10462-025-11290-y},
	volume = {58},
	year = {2025}
}

@misc{wang2025scipipllmbasedscientificpaper,
      title={SciPIP: An LLM-based Scientific Paper Idea Proposer}, 
      author={Wenxiao Wang and Lihui Gu and Liye Zhang and Yunxiang Luo and Yi Dai and Chen Shen and Liang Xie and Binbin Lin and Xiaofei He and Jieping Ye},
      year={2025},
      eprint={2410.23166},
      archivePrefix={arXiv},
      primaryClass={cs.CL},
      url={https://arxiv.org/abs/2410.23166}, 
}

@misc{lu2024aiscientistfullyautomated,
      title={The AI Scientist: Towards Fully Automated Open-Ended Scientific Discovery}, 
      author={Chris Lu and Cong Lu and Robert Tjarko Lange and Jakob Foerster and Jeff Clune and David Ha},
      year={2024},
      eprint={2408.06292},
      archivePrefix={arXiv},
      primaryClass={cs.AI},
      url={https://arxiv.org/abs/2408.06292}, 
}

@article{doi:10.1126/science.aao0185,
author = {Santo Fortunato  and Carl T. Bergstrom  and Katy Börner  and James A. Evans  and Dirk Helbing  and Staša Milojević  and Alexander M. Petersen  and Filippo Radicchi  and Roberta Sinatra  and Brian Uzzi  and Alessandro Vespignani  and Ludo Waltman  and Dashun Wang  and Albert-László Barabási },
title = {Science of science},
journal = {Science},
volume = {359},
number = {6379},
pages = {eaao0185},
year = {2018},
doi = {10.1126/science.aao0185},
URL = {https://www.science.org/doi/abs/10.1126/science.aao0185},
eprint = {https://www.science.org/doi/pdf/10.1126/science.aao0185}
}

@inproceedings{wang2019idea,
  title        = {Towards Computational Assessment of Idea Novelty},
  author       = {Wang, Kai and Dong, Boxiang and Ma, Junjie},
  booktitle    = {Proceedings of the 52nd Hawaii International Conference on System Sciences},
  year         = {2019},
  month        = {May},
  isbn         = {978-0-9981331-2-6},
  url          = {https://ssrn.com/abstract=3393611}
}

@article{SARICA2020112995,
title = {TechNet: Technology semantic network based on patent data},
journal = {Expert Systems with Applications},
volume = {142},
pages = {112995},
year = {2020},
issn = {0957-4174},
doi = {https://doi.org/10.1016/j.eswa.2019.112995},
url = {https://www.sciencedirect.com/science/article/pii/S0957417419307122},
author = {Serhad Sarica and Jianxi Luo and Kristin L. Wood}
}

@misc{gómezpérez2022artificialintelligencenaturallanguage,
      title={Artificial Intelligence and Natural Language Processing and Understanding in Space: A Methodological Framework and Four ESA Case Studies}, 
      author={José Manuel Gómez-Pérez and Andrés García-Silva and Rosemarie Leone and Mirko Albani and Moritz Fontaine and Charles Poncet and Leopold Summerer and Alessandro Donati and Ilaria Roma and Stefano Scaglioni},
      year={2022},
      eprint={2210.03640},
      archivePrefix={arXiv},
      primaryClass={cs.CL},
      url={https://arxiv.org/abs/2210.03640}, 
}

@misc{openai2025gptoss120bgptoss20bmodel,
      title={gpt-oss-120b \& gpt-oss-20b Model Card}, 
      author={OpenAI and : and Sandhini Agarwal and Lama Ahmad and Jason Ai and Sam Altman and Andy Applebaum and Edwin Arbus and Rahul K. Arora and Yu Bai and Bowen Baker and Haiming Bao and Boaz Barak and Ally Bennett and Tyler Bertao and Nivedita Brett and Eugene Brevdo and Greg Brockman and Sebastien Bubeck and Che Chang and Kai Chen and Mark Chen and Enoch Cheung and Aidan Clark and Dan Cook and Marat Dukhan and Casey Dvorak and Kevin Fives and Vlad Fomenko and Timur Garipov and Kristian Georgiev and Mia Glaese and Tarun Gogineni and Adam Goucher and Lukas Gross and Katia Gil Guzman and John Hallman and Jackie Hehir and Johannes Heidecke and Alec Helyar and Haitang Hu and Romain Huet and Jacob Huh and Saachi Jain and Zach Johnson and Chris Koch and Irina Kofman and Dominik Kundel and Jason Kwon and Volodymyr Kyrylov and Elaine Ya Le and Guillaume Leclerc and James Park Lennon and Scott Lessans and Mario Lezcano-Casado and Yuanzhi Li and Zhuohan Li and Ji Lin and Jordan Liss and Lily and Liu and Jiancheng Liu and Kevin Lu and Chris Lu and Zoran Martinovic and Lindsay McCallum and Josh McGrath and Scott McKinney and Aidan McLaughlin and Song Mei and Steve Mostovoy and Tong Mu and Gideon Myles and Alexander Neitz and Alex Nichol and Jakub Pachocki and Alex Paino and Dana Palmie and Ashley Pantuliano and Giambattista Parascandolo and Jongsoo Park and Leher Pathak and Carolina Paz and Ludovic Peran and Dmitry Pimenov and Michelle Pokrass and Elizabeth Proehl and Huida Qiu and Gaby Raila and Filippo Raso and Hongyu Ren and Kimmy Richardson and David Robinson and Bob Rotsted and Hadi Salman and Suvansh Sanjeev and Max Schwarzer and D. Sculley and Harshit Sikchi and Kendal Simon and Karan Singhal and Yang Song and Dane Stuckey and Zhiqing Sun and Philippe Tillet and Sam Toizer and Foivos Tsimpourlas and Nikhil Vyas and Eric Wallace and Xin Wang and Miles Wang and Olivia Watkins and Kevin Weil and Amy Wendling and Kevin Whinnery and Cedric Whitney and Hannah Wong and Lin Yang and Yu Yang and Michihiro Yasunaga and Kristen Ying and Wojciech Zaremba and Wenting Zhan and Cyril Zhang and Brian Zhang and Eddie Zhang and Shengjia Zhao},
      year={2025},
      eprint={2508.10925},
      archivePrefix={arXiv},
      primaryClass={cs.CL},
      url={https://arxiv.org/abs/2508.10925}, 
}

@misc{grattafiori2024llama3herdmodels,
      title={The Llama 3 Herd of Models}, 
      author={Aaron Grattafiori and Abhimanyu Dubey and Abhinav Jauhri and Abhinav Pandey and Abhishek Kadian and Ahmad Al-Dahle and Aiesha Letman and Akhil Mathur and Alan Schelten and Alex Vaughan and Amy Yang and Angela Fan and Anirudh Goyal and Anthony Hartshorn and Aobo Yang and Archi Mitra and Archie Sravankumar and Artem Korenev and Arthur Hinsvark and Arun Rao and Aston Zhang and Aurelien Rodriguez and Austen Gregerson and Ava Spataru and Baptiste Roziere and Bethany Biron and Binh Tang and Bobbie Chern and Charlotte Caucheteux and Chaya Nayak and Chloe Bi and Chris Marra and Chris McConnell and Christian Keller and Christophe Touret and Chunyang Wu and Corinne Wong and Cristian Canton Ferrer and Cyrus Nikolaidis and Damien Allonsius and Daniel Song and Danielle Pintz and Danny Livshits and Danny Wyatt and David Esiobu and Dhruv Choudhary and Dhruv Mahajan and Diego Garcia-Olano and Diego Perino and Dieuwke Hupkes and Egor Lakomkin and Ehab AlBadawy and Elina Lobanova and Emily Dinan and Eric Michael Smith and Filip Radenovic and Francisco Guzmán and Frank Zhang and Gabriel Synnaeve and Gabrielle Lee and Georgia Lewis Anderson and Govind Thattai and Graeme Nail and Gregoire Mialon and Guan Pang and Guillem Cucurell and Hailey Nguyen and Hannah Korevaar and Hu Xu and Hugo Touvron and Iliyan Zarov and Imanol Arrieta Ibarra and Isabel Kloumann and Ishan Misra and Ivan Evtimov and Jack Zhang and Jade Copet and Jaewon Lee and Jan Geffert and Jana Vranes and Jason Park and Jay Mahadeokar and Jeet Shah and Jelmer van der Linde and Jennifer Billock and Jenny Hong and Jenya Lee and Jeremy Fu and Jianfeng Chi and Jianyu Huang and Jiawen Liu and Jie Wang and Jiecao Yu and Joanna Bitton and Joe Spisak and Jongsoo Park and Joseph Rocca and Joshua Johnstun and Joshua Saxe and Junteng Jia and Kalyan Vasuden Alwala and Karthik Prasad and Kartikeya Upasani and Kate Plawiak and Ke Li and Kenneth Heafield and Kevin Stone and Khalid El-Arini and Krithika Iyer and Kshitiz Malik and Kuenley Chiu and Kunal Bhalla and Kushal Lakhotia and Lauren Rantala-Yeary and Laurens van der Maaten and Lawrence Chen and Liang Tan and Liz Jenkins and Louis Martin and Lovish Madaan and Lubo Malo and Lukas Blecher and Lukas Landzaat and Luke de Oliveira and Madeline Muzzi and Mahesh Pasupuleti and Mannat Singh and Manohar Paluri and Marcin Kardas and Maria Tsimpoukelli and Mathew Oldham and Mathieu Rita and Maya Pavlova and Melanie Kambadur and Mike Lewis and Min Si and Mitesh Kumar Singh and Mona Hassan and Naman Goyal and Narjes Torabi and Nikolay Bashlykov and Nikolay Bogoychev and Niladri Chatterji and Ning Zhang and Olivier Duchenne and Onur Çelebi and Patrick Alrassy and Pengchuan Zhang and Pengwei Li and Petar Vasic and Peter Weng and Prajjwal Bhargava and Pratik Dubal and Praveen Krishnan and Punit Singh Koura and Puxin Xu and Qing He and Qingxiao Dong and Ragavan Srinivasan and Raj Ganapathy and Ramon Calderer and Ricardo Silveira Cabral and Robert Stojnic and Roberta Raileanu and Rohan Maheswari and Rohit Girdhar and Rohit Patel and Romain Sauvestre and Ronnie Polidoro and Roshan Sumbaly and Ross Taylor and Ruan Silva and Rui Hou and Rui Wang and Saghar Hosseini and Sahana Chennabasappa and Sanjay Singh and Sean Bell and Seohyun Sonia Kim and Sergey Edunov and Shaoliang Nie and Sharan Narang and Sharath Raparthy and Sheng Shen and Shengye Wan and Shruti Bhosale and Shun Zhang and Simon Vandenhende and Soumya Batra and Spencer Whitman and Sten Sootla and Stephane Collot and Suchin Gururangan and Sydney Borodinsky and Tamar Herman and Tara Fowler and Tarek Sheasha and Thomas Georgiou and Thomas Scialom and Tobias Speckbacher and Todor Mihaylov and Tong Xiao and Ujjwal Karn and Vedanuj Goswami and Vibhor Gupta and Vignesh Ramanathan and Viktor Kerkez and Vincent Gonguet and Virginie Do and Vish Vogeti and Vítor Albiero and Vladan Petrovic and Weiwei Chu and Wenhan Xiong and Wenyin Fu and Whitney Meers and Xavier Martinet and Xiaodong Wang and Xiaofang Wang and Xiaoqing Ellen Tan and Xide Xia and Xinfeng Xie and Xuchao Jia and Xuewei Wang and Yaelle Goldschlag and Yashesh Gaur and Yasmine Babaei and Yi Wen and Yiwen Song and Yuchen Zhang and Yue Li and Yuning Mao and Zacharie Delpierre Coudert and Zheng Yan and Zhengxing Chen and Zoe Papakipos and Aaditya Singh and Aayushi Srivastava and Abha Jain and Adam Kelsey and Adam Shajnfeld and Adithya Gangidi and Adolfo Victoria and Ahuva Goldstand and Ajay Menon and Ajay Sharma and Alex Boesenberg and Alexei Baevski and Allie Feinstein and Amanda Kallet and Amit Sangani and Amos Teo and Anam Yunus and Andrei Lupu and Andres Alvarado and Andrew Caples and Andrew Gu and Andrew Ho and Andrew Poulton and Andrew Ryan and Ankit Ramchandani and Annie Dong and Annie Franco and Anuj Goyal and Aparajita Saraf and Arkabandhu Chowdhury and Ashley Gabriel and Ashwin Bharambe and Assaf Eisenman and Azadeh Yazdan and Beau James and Ben Maurer and Benjamin Leonhardi and Bernie Huang and Beth Loyd and Beto De Paola and Bhargavi Paranjape and Bing Liu and Bo Wu and Boyu Ni and Braden Hancock and Bram Wasti and Brandon Spence and Brani Stojkovic and Brian Gamido and Britt Montalvo and Carl Parker and Carly Burton and Catalina Mejia and Ce Liu and Changhan Wang and Changkyu Kim and Chao Zhou and Chester Hu and Ching-Hsiang Chu and Chris Cai and Chris Tindal and Christoph Feichtenhofer and Cynthia Gao and Damon Civin and Dana Beaty and Daniel Kreymer and Daniel Li and David Adkins and David Xu and Davide Testuggine and Delia David and Devi Parikh and Diana Liskovich and Didem Foss and Dingkang Wang and Duc Le and Dustin Holland and Edward Dowling and Eissa Jamil and Elaine Montgomery and Eleonora Presani and Emily Hahn and Emily Wood and Eric-Tuan Le and Erik Brinkman and Esteban Arcaute and Evan Dunbar and Evan Smothers and Fei Sun and Felix Kreuk and Feng Tian and Filippos Kokkinos and Firat Ozgenel and Francesco Caggioni and Frank Kanayet and Frank Seide and Gabriela Medina Florez and Gabriella Schwarz and Gada Badeer and Georgia Swee and Gil Halpern and Grant Herman and Grigory Sizov and Guangyi and Zhang and Guna Lakshminarayanan and Hakan Inan and Hamid Shojanazeri and Han Zou and Hannah Wang and Hanwen Zha and Haroun Habeeb and Harrison Rudolph and Helen Suk and Henry Aspegren and Hunter Goldman and Hongyuan Zhan and Ibrahim Damlaj and Igor Molybog and Igor Tufanov and Ilias Leontiadis and Irina-Elena Veliche and Itai Gat and Jake Weissman and James Geboski and James Kohli and Janice Lam and Japhet Asher and Jean-Baptiste Gaya and Jeff Marcus and Jeff Tang and Jennifer Chan and Jenny Zhen and Jeremy Reizenstein and Jeremy Teboul and Jessica Zhong and Jian Jin and Jingyi Yang and Joe Cummings and Jon Carvill and Jon Shepard and Jonathan McPhie and Jonathan Torres and Josh Ginsburg and Junjie Wang and Kai Wu and Kam Hou U and Karan Saxena and Kartikay Khandelwal and Katayoun Zand and Kathy Matosich and Kaushik Veeraraghavan and Kelly Michelena and Keqian Li and Kiran Jagadeesh and Kun Huang and Kunal Chawla and Kyle Huang and Lailin Chen and Lakshya Garg and Lavender A and Leandro Silva and Lee Bell and Lei Zhang and Liangpeng Guo and Licheng Yu and Liron Moshkovich and Luca Wehrstedt and Madian Khabsa and Manav Avalani and Manish Bhatt and Martynas Mankus and Matan Hasson and Matthew Lennie and Matthias Reso and Maxim Groshev and Maxim Naumov and Maya Lathi and Meghan Keneally and Miao Liu and Michael L. Seltzer and Michal Valko and Michelle Restrepo and Mihir Patel and Mik Vyatskov and Mikayel Samvelyan and Mike Clark and Mike Macey and Mike Wang and Miquel Jubert Hermoso and Mo Metanat and Mohammad Rastegari and Munish Bansal and Nandhini Santhanam and Natascha Parks and Natasha White and Navyata Bawa and Nayan Singhal and Nick Egebo and Nicolas Usunier and Nikhil Mehta and Nikolay Pavlovich Laptev and Ning Dong and Norman Cheng and Oleg Chernoguz and Olivia Hart and Omkar Salpekar and Ozlem Kalinli and Parkin Kent and Parth Parekh and Paul Saab and Pavan Balaji and Pedro Rittner and Philip Bontrager and Pierre Roux and Piotr Dollar and Polina Zvyagina and Prashant Ratanchandani and Pritish Yuvraj and Qian Liang and Rachad Alao and Rachel Rodriguez and Rafi Ayub and Raghotham Murthy and Raghu Nayani and Rahul Mitra and Rangaprabhu Parthasarathy and Raymond Li and Rebekkah Hogan and Robin Battey and Rocky Wang and Russ Howes and Ruty Rinott and Sachin Mehta and Sachin Siby and Sai Jayesh Bondu and Samyak Datta and Sara Chugh and Sara Hunt and Sargun Dhillon and Sasha Sidorov and Satadru Pan and Saurabh Mahajan and Saurabh Verma and Seiji Yamamoto and Sharadh Ramaswamy and Shaun Lindsay and Shaun Lindsay and Sheng Feng and Shenghao Lin and Shengxin Cindy Zha and Shishir Patil and Shiva Shankar and Shuqiang Zhang and Shuqiang Zhang and Sinong Wang and Sneha Agarwal and Soji Sajuyigbe and Soumith Chintala and Stephanie Max and Stephen Chen and Steve Kehoe and Steve Satterfield and Sudarshan Govindaprasad and Sumit Gupta and Summer Deng and Sungmin Cho and Sunny Virk and Suraj Subramanian and Sy Choudhury and Sydney Goldman and Tal Remez and Tamar Glaser and Tamara Best and Thilo Koehler and Thomas Robinson and Tianhe Li and Tianjun Zhang and Tim Matthews and Timothy Chou and Tzook Shaked and Varun Vontimitta and Victoria Ajayi and Victoria Montanez and Vijai Mohan and Vinay Satish Kumar and Vishal Mangla and Vlad Ionescu and Vlad Poenaru and Vlad Tiberiu Mihailescu and Vladimir Ivanov and Wei Li and Wenchen Wang and Wenwen Jiang and Wes Bouaziz and Will Constable and Xiaocheng Tang and Xiaojian Wu and Xiaolan Wang and Xilun Wu and Xinbo Gao and Yaniv Kleinman and Yanjun Chen and Ye Hu and Ye Jia and Ye Qi and Yenda Li and Yilin Zhang and Ying Zhang and Yossi Adi and Youngjin Nam and Yu and Wang and Yu Zhao and Yuchen Hao and Yundi Qian and Yunlu Li and Yuzi He and Zach Rait and Zachary DeVito and Zef Rosnbrick and Zhaoduo Wen and Zhenyu Yang and Zhiwei Zhao and Zhiyu Ma},
      year={2024},
      eprint={2407.21783},
      archivePrefix={arXiv},
      primaryClass={cs.AI},
      url={https://arxiv.org/abs/2407.21783}, 
}

@misc{OpenAI-gpt-5.4, 
url={https://deploymentsafety.openai.com/gpt-5-4-thinking/gpt-5-4-thinking.pdf}, 
title={GPT-5.4 Thinking System Card}, 
publisher={OpenAI}, 
author={OpenAI},
year={2026}, 
month={Mar},
Note={Accessed: 2026-05-20}
}

@inproceedings{zhang2025noveltybench,
title={NoveltyBench: Evaluating Creativity and Diversity in Language Models},
author={Yiming Zhang and Harshita Diddee and Susan Holm and Hanchen Liu and Xinyue Liu and Vinay Samuel and Barry Wang and Daphne Ippolito},
booktitle={Second Conference on Language Modeling},
year={2025},
url={https://openreview.net/forum?id=XZm1ekzERf}
}

@inproceedings{tang2025airesearcher,
    title={{AI}-Researcher: Autonomous Scientific Innovation},
    author={Jiabin Tang and Lianghao Xia and Zhonghang Li and Chao Huang},
    booktitle={The Thirty-ninth Annual Conference on Neural Information Processing Systems},
    year={2025},
    url={https://openreview.net/forum?id=kQWyOYUAC4}
}

@inproceedings{NEURIPS2024_b1b16c4b,
 author = {Li, Zichao and Cao, Yanshuai and Cheung, Jackie C.K.},
 booktitle = {Advances in Neural Information Processing Systems},
 doi = {10.52202/079017-3110},
 editor = {A. Globerson and L. Mackey and D. Belgrave and A. Fan and U. Paquet and J. Tomczak and C. Zhang},
 pages = {98009--98032},
 publisher = {Curran Associates, Inc.},
 title = {Do LLMs Build World Representations? Probing Through the Lens of State Abstraction},
 url = {https://proceedings.neurips.cc/paper_files/paper/2024/file/b1b16c4b875eb84d3585cb70d23970ca-Paper-Conference.pdf},
 volume = {37},
 year = {2024}
}

@article{10.1162/TACL.a.48,
    author = {Valois, Pedro H. V. and Souza, Lincon S. and Shimomoto, Erica K. and Fukui, Kazuhiro},
    title = {Frame Representation Hypothesis: Multi-Token LLM Interpretability and Concept-Guided Text Generation},
    journal = {Transactions of the Association for Computational Linguistics},
    volume = {13},
    pages = {1436-1458},
    year = {2025},
    month = {10},
    issn = {2307-387X},
    doi = {10.1162/TACL.a.48},
    url = {https://doi.org/10.1162/TACL.a.48},
    eprint = {https://direct.mit.edu/tacl/article-pdf/doi/10.1162/TACL.a.48/2561673/tacl.a.48.pdf},
}

@misc{zur2025languagemodelsawareroad,
      title={Are language models aware of the road not taken? Token-level uncertainty and hidden state dynamics}, 
      author={Amir Zur and Atticus Geiger and Ekdeep Singh Lubana and Eric Bigelow},
      year={2025},
      eprint={2511.04527},
      archivePrefix={arXiv},
      primaryClass={cs.CL},
      url={https://arxiv.org/abs/2511.04527}, 
}

@inproceedings{marks2024the,
title={The Geometry of Truth: Emergent Linear Structure in Large Language Model Representations of True/False Datasets},
author={Samuel Marks and Max Tegmark},
booktitle={First Conference on Language Modeling},
year={2024},
url={https://openreview.net/forum?id=aajyHYjjsk}
}

@inproceedings{gurnee2024language,
title={Language Models Represent Space and Time},
author={Wes Gurnee and Max Tegmark},
booktitle={The Twelfth International Conference on Learning Representations},
year={2024},
url={https://openreview.net/forum?id=jE8xbmvFin}
}

@inproceedings{li2023emergent,
title={Emergent World Representations: Exploring a Sequence Model Trained on a Synthetic Task},
author={Kenneth Li and Aspen K Hopkins and David Bau and Fernanda Vi{\'e}gas and Hanspeter Pfister and Martin Wattenberg},
booktitle={The Eleventh International Conference on Learning Representations },
year={2023},
url={https://openreview.net/forum?id=DeG07_TcZvT}
}

@misc{anthropicIntroducingClaudeOpus,
	author = {Anthropic},
	title = {{I}ntroducing {C}laude {O}pus 4.5 --- anthropic.com},
	howpublished = {\url{https://www.anthropic.com/news/claude-opus-4-5}},
	year = {2025},
	note = {[Accessed 05-01-2026]},
}

@misc{gemini-3,
	author = {Google},
	title = {{A} new era of intelligence with {G}emini 3 --- blog.google},
	howpublished = {\url{https://blog.google/products/gemini/gemini-3/#note-from-ceo}},
	year = {2025},
	note = {[Accessed 05-01-2026]},
}

@misc{anthropicIntroducingClaudeSonnet,
	author = {Anthropic},
	title = {{I}ntroducing {C}laude {S}onnet 4.5 --- anthropic.com},
	howpublished = {\url{https://www.anthropic.com/news/claude-sonnet-4-5}},
	year = {2025},
	note = {[Accessed 05-01-2026]},
}

@inproceedings{hu2022lora,
title={Lo{RA}: Low-Rank Adaptation of Large Language Models},
author={Edward J Hu and yelong shen and Phillip Wallis and Zeyuan Allen-Zhu and Yuanzhi Li and Shean Wang and Lu Wang and Weizhu Chen},
booktitle={International Conference on Learning Representations},
year={2022},
url={https://openreview.net/forum?id=nZeVKeeFYf9}
}

@misc{gemmateam2025gemma3technicalreport,
      title={Gemma 3 Technical Report}, 
      author={Google and Aishwarya Kamath and Johan Ferret and Shreya Pathak and Nino Vieillard and Ramona Merhej and Sarah Perrin and Tatiana Matejovicova and Alexandre Ramé and Morgane Rivière and Louis Rouillard and Thomas Mesnard and Geoffrey Cideron and Jean-bastien Grill and Sabela Ramos and Edouard Yvinec and Michelle Casbon and Etienne Pot and Ivo Penchev and Gaël Liu and Francesco Visin and Kathleen Kenealy and Lucas Beyer and Xiaohai Zhai and Anton Tsitsulin and Robert Busa-Fekete and Alex Feng and Noveen Sachdeva and Benjamin Coleman and Yi Gao and Basil Mustafa and Iain Barr and Emilio Parisotto and David Tian and Matan Eyal and Colin Cherry and Jan-Thorsten Peter and Danila Sinopalnikov and Surya Bhupatiraju and Rishabh Agarwal and Mehran Kazemi and Dan Malkin and Ravin Kumar and David Vilar and Idan Brusilovsky and Jiaming Luo and Andreas Steiner and Abe Friesen and Abhanshu Sharma and Abheesht Sharma and Adi Mayrav Gilady and Adrian Goedeckemeyer and Alaa Saade and Alex Feng and Alexander Kolesnikov and Alexei Bendebury and Alvin Abdagic and Amit Vadi and András György and André Susano Pinto and Anil Das and Ankur Bapna and Antoine Miech and Antoine Yang and Antonia Paterson and Ashish Shenoy and Ayan Chakrabarti and Bilal Piot and Bo Wu and Bobak Shahriari and Bryce Petrini and Charlie Chen and Charline Le Lan and Christopher A. Choquette-Choo and CJ Carey and Cormac Brick and Daniel Deutsch and Danielle Eisenbud and Dee Cattle and Derek Cheng and Dimitris Paparas and Divyashree Shivakumar Sreepathihalli and Doug Reid and Dustin Tran and Dustin Zelle and Eric Noland and Erwin Huizenga and Eugene Kharitonov and Frederick Liu and Gagik Amirkhanyan and Glenn Cameron and Hadi Hashemi and Hanna Klimczak-Plucińska and Harman Singh and Harsh Mehta and Harshal Tushar Lehri and Hussein Hazimeh and Ian Ballantyne and Idan Szpektor and Ivan Nardini and Jean Pouget-Abadie and Jetha Chan and Joe Stanton and John Wieting and Jonathan Lai and Jordi Orbay and Joseph Fernandez and Josh Newlan and Ju-yeong Ji and Jyotinder Singh and Kat Black and Kathy Yu and Kevin Hui and Kiran Vodrahalli and Klaus Greff and Linhai Qiu and Marcella Valentine and Marina Coelho and Marvin Ritter and Matt Hoffman and Matthew Watson and Mayank Chaturvedi and Michael Moynihan and Min Ma and Nabila Babar and Natasha Noy and Nathan Byrd and Nick Roy and Nikola Momchev and Nilay Chauhan and Noveen Sachdeva and Oskar Bunyan and Pankil Botarda and Paul Caron and Paul Kishan Rubenstein and Phil Culliton and Philipp Schmid and Pier Giuseppe Sessa and Pingmei Xu and Piotr Stanczyk and Pouya Tafti and Rakesh Shivanna and Renjie Wu and Renke Pan and Reza Rokni and Rob Willoughby and Rohith Vallu and Ryan Mullins and Sammy Jerome and Sara Smoot and Sertan Girgin and Shariq Iqbal and Shashir Reddy and Shruti Sheth and Siim Põder and Sijal Bhatnagar and Sindhu Raghuram Panyam and Sivan Eiger and Susan Zhang and Tianqi Liu and Trevor Yacovone and Tyler Liechty and Uday Kalra and Utku Evci and Vedant Misra and Vincent Roseberry and Vlad Feinberg and Vlad Kolesnikov and Woohyun Han and Woosuk Kwon and Xi Chen and Yinlam Chow and Yuvein Zhu and Zichuan Wei and Zoltan Egyed and Victor Cotruta and Minh Giang and Phoebe Kirk and Anand Rao and Kat Black and Nabila Babar and Jessica Lo and Erica Moreira and Luiz Gustavo Martins and Omar Sanseviero and Lucas Gonzalez and Zach Gleicher and Tris Warkentin and Vahab Mirrokni and Evan Senter and Eli Collins and Joelle Barral and Zoubin Ghahramani and Raia Hadsell and Yossi Matias and D. Sculley and Slav Petrov and Noah Fiedel and Noam Shazeer and Oriol Vinyals and Jeff Dean and Demis Hassabis and Koray Kavukcuoglu and Clement Farabet and Elena Buchatskaya and Jean-Baptiste Alayrac and Rohan Anil and Dmitry and Lepikhin and Sebastian Borgeaud and Olivier Bachem and Armand Joulin and Alek Andreev and Cassidy Hardin and Robert Dadashi and Léonard Hussenot},
      year={2025},
      eprint={2503.19786},
      archivePrefix={arXiv},
      primaryClass={cs.CL},
      url={https://arxiv.org/abs/2503.19786}, 
}

@misc{yang2025qwen3technicalreport,
      title={Qwen3 Technical Report}, 
      author={An Yang and Anfeng Li and Baosong Yang and Beichen Zhang and Binyuan Hui and Bo Zheng and Bowen Yu and Chang Gao and Chengen Huang and Chenxu Lv and Chujie Zheng and Dayiheng Liu and Fan Zhou and Fei Huang and Feng Hu and Hao Ge and Haoran Wei and Huan Lin and Jialong Tang and Jian Yang and Jianhong Tu and Jianwei Zhang and Jianxin Yang and Jiaxi Yang and Jing Zhou and Jingren Zhou and Junyang Lin and Kai Dang and Keqin Bao and Kexin Yang and Le Yu and Lianghao Deng and Mei Li and Mingfeng Xue and Mingze Li and Pei Zhang and Peng Wang and Qin Zhu and Rui Men and Ruize Gao and Shixuan Liu and Shuang Luo and Tianhao Li and Tianyi Tang and Wenbiao Yin and Xingzhang Ren and Xinyu Wang and Xinyu Zhang and Xuancheng Ren and Yang Fan and Yang Su and Yichang Zhang and Yinger Zhang and Yu Wan and Yuqiong Liu and Zekun Wang and Zeyu Cui and Zhenru Zhang and Zhipeng Zhou and Zihan Qiu},
      year={2025},
      eprint={2505.09388},
      archivePrefix={arXiv},
      primaryClass={cs.CL},
      url={https://arxiv.org/abs/2505.09388}, 
}

@misc{liu2025harnessinglargelanguagemodels,
      title={Harnessing Large Language Models for Scientific Novelty Detection}, 
      author={Yan Liu and Zonglin Yang and Soujanya Poria and Thanh-Son Nguyen and Erik Cambria},
      year={2025},
      eprint={2505.24615},
      archivePrefix={arXiv},
      primaryClass={cs.CL},
      url={https://arxiv.org/abs/2505.24615}, 
}

@inproceedings{feng2025grapheval,
title={GraphEval: A Lightweight Graph-Based {LLM} Framework for Idea Evaluation},
author={Tao Feng and Yihang Sun and Jiaxuan You},
booktitle={The Thirteenth International Conference on Learning Representations},
year={2025},
url={https://openreview.net/forum?id=5RUM1aIdok}
}

@article{doi:10.1126/science.1240474,
author = {Brian Uzzi  and Satyam Mukherjee  and Michael Stringer  and Ben Jones },
title = {Atypical Combinations and Scientific Impact},
journal = {Science},
volume = {342},
number = {6157},
pages = {468-472},
year = {2013},
doi = {10.1126/science.1240474},
URL = {https://www.science.org/doi/abs/10.1126/science.1240474},
eprint = {https://www.science.org/doi/pdf/10.1126/science.1240474}}

@article{WANG20171416,
title = {Bias against novelty in science: A cautionary tale for users of bibliometric indicators},
journal = {Research Policy},
volume = {46},
number = {8},
pages = {1416-1436},
year = {2017},
issn = {0048-7333},
doi = {https://doi.org/10.1016/j.respol.2017.06.006},
url = {https://www.sciencedirect.com/science/article/pii/S0048733317301038},
author = {Jian Wang and Reinhilde Veugelers and Paula Stephan}
}

@article{https://doi.org/10.1002/asi.70005,
author = {Wu, Wenqing and Zhang, Chengzhi and Zhao, Yi},
title = {Automated novelty evaluation of academic paper: A collaborative approach integrating human and large language model knowledge},
journal = {Journal of the Association for Information Science and Technology},
volume = {76},
number = {11},
pages = {1452-1469},
doi = {https://doi.org/10.1002/asi.70005},
url = {https://asistdl.onlinelibrary.wiley.com/doi/abs/10.1002/asi.70005},
eprint = {https://asistdl.onlinelibrary.wiley.com/doi/pdf/10.1002/asi.70005},
year = {2025}
}

@misc{li2024chainideasrevolutionizingresearch,
      title={Chain of Ideas: Revolutionizing Research Via Novel Idea Development with LLM Agents}, 
      author={Long Li and Weiwen Xu and Jiayan Guo and Ruochen Zhao and Xingxuan Li and Yuqian Yuan and Boqiang Zhang and Yuming Jiang and Yifei Xin and Ronghao Dang and Deli Zhao and Yu Rong and Tian Feng and Lidong Bing},
      year={2024},
      eprint={2410.13185},
      archivePrefix={arXiv},
      primaryClass={cs.AI},
      url={https://arxiv.org/abs/2410.13185}, 
}

@inproceedings{schopf-etal-2026-is,
  title = {Is This Idea Novel? An Automated Benchmark for Judgment of Research Ideas},
  author = {Schopf, Tim and Färber, Michael},
  booktitle = {Proceedings of the Fifteenth Language Resources and Evaluation Conference (LREC 2026)},
  month = {May},
  year = {2026},
  pages = {4716--4727},
  address = {Palma, Mallorca, Spain},
  publisher = {European Language Resources Association (ELRA)},
  editor = {Piperidis, Stelios and Bel, Núria and van den Heuvel, Henk and Ide, Nancy and Krek, Simon and Toral, Antonio},
  doi = {10.63317/4c3gy3f7epnj}
}

@misc{singh2026openaigpt5card,
      title={OpenAI GPT-5 System Card}, 
      author={Aaditya Singh and Adam Fry and Adam Perelman and Adam Tart and Adi Ganesh and Ahmed El-Kishky and Aidan McLaughlin and Aiden Low and AJ Ostrow and Akhila Ananthram and Akshay Nathan and Alan Luo and Alec Helyar and Aleksander Madry and Aleksandr Efremov and Aleksandra Spyra and Alex Baker-Whitcomb and Alex Beutel and Alex Karpenko and Alex Makelov and Alex Neitz and Alex Wei and Alexandra Barr and Alexandre Kirchmeyer and Alexey Ivanov and Alexi Christakis and Alistair Gillespie and Allison Tam and Ally Bennett and Alvin Wan and Alyssa Huang and Amy McDonald Sandjideh and Amy Yang and Ananya Kumar and Andre Saraiva and Andrea Vallone and Andrei Gheorghe and Andres Garcia Garcia and Andrew Braunstein and Andrew Liu and Andrew Schmidt and Andrey Mereskin and Andrey Mishchenko and Andy Applebaum and Andy Rogerson and Ann Rajan and Annie Wei and Anoop Kotha and Anubha Srivastava and Anushree Agrawal and Arun Vijayvergiya and Ashley Tyra and Ashvin Nair and Avi Nayak and Ben Eggers and Bessie Ji and Beth Hoover and Bill Chen and Blair Chen and Boaz Barak and Borys Minaiev and Botao Hao and Bowen Baker and Brad Lightcap and Brandon McKinzie and Brandon Wang and Brendan Quinn and Brian Fioca and Brian Hsu and Brian Yang and Brian Yu and Brian Zhang and Brittany Brenner and Callie Riggins Zetino and Cameron Raymond and Camillo Lugaresi and Carolina Paz and Cary Hudson and Cedric Whitney and Chak Li and Charles Chen and Charlotte Cole and Chelsea Voss and Chen Ding and Chen Shen and Chengdu Huang and Chris Colby and Chris Hallacy and Chris Koch and Chris Lu and Christina Kaplan and Christina Kim and CJ Minott-Henriques and Cliff Frey and Cody Yu and Coley Czarnecki and Colin Reid and Colin Wei and Cory Decareaux and Cristina Scheau and Cyril Zhang and Cyrus Forbes and Da Tang and Dakota Goldberg and Dan Roberts and Dana Palmie and Daniel Kappler and Daniel Levine and Daniel Wright and Dave Leo and David Lin and David Robinson and Declan Grabb and Derek Chen and Derek Lim and Derek Salama and Dibya Bhattacharjee and Dimitris Tsipras and Dinghua Li and Dingli Yu and DJ Strouse and Drew Williams and Dylan Hunn and Ed Bayes and Edwin Arbus and Ekin Akyurek and Elaine Ya Le and Elana Widmann and Eli Yani and Elizabeth Proehl and Enis Sert and Enoch Cheung and Eri Schwartz and Eric Han and Eric Jiang and Eric Mitchell and Eric Sigler and Eric Wallace and Erik Ritter and Erin Kavanaugh and Evan Mays and Evgenii Nikishin and Fangyuan Li and Felipe Petroski Such and Filipe de Avila Belbute Peres and Filippo Raso and Florent Bekerman and Foivos Tsimpourlas and Fotis Chantzis and Francis Song and Francis Zhang and Gaby Raila and Garrett McGrath and Gary Briggs and Gary Yang and Giambattista Parascandolo and Gildas Chabot and Grace Kim and Grace Zhao and Gregory Valiant and Guillaume Leclerc and Hadi Salman and Hanson Wang and Hao Sheng and Haoming Jiang and Haoyu Wang and Haozhun Jin and Harshit Sikchi and Heather Schmidt and Henry Aspegren and Honglin Chen and Huida Qiu and Hunter Lightman and Ian Covert and Ian Kivlichan and Ian Silber and Ian Sohl and Ibrahim Hammoud and Ignasi Clavera and Ikai Lan and Ilge Akkaya and Ilya Kostrikov and Irina Kofman and Isak Etinger and Ishaan Singal and Jackie Hehir and Jacob Huh and Jacqueline Pan and Jake Wilczynski and Jakub Pachocki and James Lee and James Quinn and Jamie Kiros and Janvi Kalra and Jasmyn Samaroo and Jason Wang and Jason Wolfe and Jay Chen and Jay Wang and Jean Harb and Jeffrey Han and Jeffrey Wang and Jennifer Zhao and Jeremy Chen and Jerene Yang and Jerry Tworek and Jesse Chand and Jessica Landon and Jessica Liang and Ji Lin and Jiancheng Liu and Jianfeng Wang and Jie Tang and Jihan Yin and Joanne Jang and Joel Morris and Joey Flynn and Johannes Ferstad and Johannes Heidecke and John Fishbein and John Hallman and Jonah Grant and Jonathan Chien and Jonathan Gordon and Jongsoo Park and Jordan Liss and Jos Kraaijeveld and Joseph Guay and Joseph Mo and Josh Lawson and Josh McGrath and Joshua Vendrow and Joy Jiao and Julian Lee and Julie Steele and Julie Wang and Junhua Mao and Kai Chen and Kai Hayashi and Kai Xiao and Kamyar Salahi and Kan Wu and Karan Sekhri and Karan Sharma and Karan Singhal and Karen Li and Kenny Nguyen and Keren Gu-Lemberg and Kevin King and Kevin Liu and Kevin Stone and Kevin Yu and Kristen Ying and Kristian Georgiev and Kristie Lim and Kushal Tirumala and Kyle Miller and Lama Ahmad and Larry Lv and Laura Clare and Laurance Fauconnet and Lauren Itow and Lauren Yang and Laurentia Romaniuk and Leah Anise and Lee Byron and Leher Pathak and Leon Maksin and Leyan Lo and Leyton Ho and Li Jing and Liang Wu and Liang Xiong and Lien Mamitsuka and Lin Yang and Lindsay McCallum and Lindsey Held and Liz Bourgeois and Logan Engstrom and Lorenz Kuhn and Louis Feuvrier and Lu Zhang and Lucas Switzer and Lukas Kondraciuk and Lukasz Kaiser and Manas Joglekar and Mandeep Singh and Mandip Shah and Manuka Stratta and Marcus Williams and Mark Chen and Mark Sun and Marselus Cayton and Martin Li and Marvin Zhang and Marwan Aljubeh and Matt Nichols and Matthew Haines and Max Schwarzer and Mayank Gupta and Meghan Shah and Melody Y. Guan and Melody Huang and Meng Dong and Mengqing Wang and Mia Glaese and Micah Carroll and Michael Lampe and Michael Malek and Michael Sharman and Michael Zhang and Michele Wang and Michelle Pokrass and Mihai Florian and Mikhail Pavlov and Miles Wang and Ming Chen and Mingxuan Wang and Minnia Feng and Mo Bavarian and Molly Lin and Moose Abdool and Mostafa Rohaninejad and Nacho Soto and Natalie Staudacher and Natan LaFontaine and Nathan Marwell and Nelson Liu and Nick Preston and Nick Turley and Nicklas Ansman and Nicole Blades and Nikil Pancha and Nikita Mikhaylin and Niko Felix and Nikunj Handa and Nishant Rai and Nitish Keskar and Noam Brown and Ofir Nachum and Oleg Boiko and Oleg Murk and Olivia Watkins and Oona Gleeson and Pamela Mishkin and Patryk Lesiewicz and Paul Baltescu and Pavel Belov and Peter Zhokhov and Philip Pronin and Phillip Guo and Phoebe Thacker and Qi Liu and Qiming Yuan and Qinghua Liu and Rachel Dias and Rachel Puckett and Rahul Arora and Ravi Teja Mullapudi and Raz Gaon and Reah Miyara and Rennie Song and Rishabh Aggarwal and RJ Marsan and Robel Yemiru and Robert Xiong and Rohan Kshirsagar and Rohan Nuttall and Roman Tsiupa and Ronen Eldan and Rose Wang and Roshan James and Roy Ziv and Rui Shu and Ruslan Nigmatullin and Saachi Jain and Saam Talaie and Sam Altman and Sam Arnesen and Sam Toizer and Sam Toyer and Samuel Miserendino and Sandhini Agarwal and Sarah Yoo and Savannah Heon and Scott Ethersmith and Sean Grove and Sean Taylor and Sebastien Bubeck and Sever Banesiu and Shaokyi Amdo and Shengjia Zhao and Sherwin Wu and Shibani Santurkar and Shiyu Zhao and Shraman Ray Chaudhuri and Shreyas Krishnaswamy and Shuaiqi and Xia and Shuyang Cheng and Shyamal Anadkat and Simón Posada Fishman and Simon Tobin and Siyuan Fu and Somay Jain and Song Mei and Sonya Egoian and Spencer Kim and Spug Golden and SQ Mah and Steph Lin and Stephen Imm and Steve Sharpe and Steve Yadlowsky and Sulman Choudhry and Sungwon Eum and Suvansh Sanjeev and Tabarak Khan and Tal Stramer and Tao Wang and Tao Xin and Tarun Gogineni and Taya Christianson and Ted Sanders and Tejal Patwardhan and Thomas Degry and Thomas Shadwell and Tianfu Fu and Tianshi Gao and Timur Garipov and Tina Sriskandarajah and Toki Sherbakov and Tomek Korbak and Tomer Kaftan and Tomo Hiratsuka and Tongzhou Wang and Tony Song and Tony Zhao and Troy Peterson and Val Kharitonov and Victoria Chernova and Vineet Kosaraju and Vishal Kuo and Vitchyr Pong and Vivek Verma and Vlad Petrov and Wanning Jiang and Weixing Zhang and Wenda Zhou and Wenlei Xie and Wenting Zhan and Wes McCabe and Will DePue and Will Ellsworth and Wulfie Bain and Wyatt Thompson and Xiangning Chen and Xiangyu Qi and Xin Xiang and Xinwei Shi and Yann Dubois and Yaodong Yu and Yara Khakbaz and Yifan Wu and Yilei Qian and Yin Tat Lee and Yinbo Chen and Yizhen Zhang and Yizhong Xiong and Yonglong Tian and Young Cha and Yu Bai and Yu Yang and Yuan Yuan and Yuanzhi Li and Yufeng Zhang and Yuguang Yang and Yujia Jin and Yun Jiang and Yunyun Wang and Yushi Wang and Yutian Liu and Zach Stubenvoll and Zehao Dou and Zheng Wu and Zhigang Wang},
      year={2026},
      eprint={2601.03267},
      archivePrefix={arXiv},
      primaryClass={cs.CL},
      url={https://arxiv.org/abs/2601.03267}, 
}

@Article{Lu2026,
author={Lu, Chris
and Lu, Cong
and Lange, Robert Tjarko
and Yamada, Yutaro
and Hu, Shengran
and Foerster, Jakob
and Ha, David
and Clune, Jeff},
title={Towards end-to-end automation of AI research},
journal={Nature},
year={2026},
month={Mar},
day={01},
volume={651},
number={8107},
pages={914-919},
issn={1476-4687},
doi={10.1038/s41586-026-10265-5},
url={https://doi.org/10.1038/s41586-026-10265-5}
}

@Article{Gottweis2026,
author={Gottweis, Juraj
and Weng, Wei-Hung
and Daryin, Alexander
and Tu, Tao
and Sirkovic, Petar
and Myaskovsky, Artiom
and Glowaty, Grzegorz
and Weissenberger, Felix
and Orlandi, Alessio
and Popovici, Dan
and Palepu, Anil
and Rong, Keran
and Tanno, Ryutaro
and Saab, Khaled
and Zhang, Fan
and Blum, Jacob
and Carroll, Andrew
and Kulkarni, Kavita
and Toma{\v{s}}ev, Nenad
and Zverinski, Dina
and Rendulic, Ivor
and Vedadi, Elahe
and Hasler, Florian
and Rimanic, Luka
and Boia, Marina
and Budiselic, Ivan
and Feinstein, Ben
and Bellaiche, Mathias
and Sheffer, Tom
and Freyberg, Jan
and Ratcliff, Jeremy
and Bertolli, Ottavia
and Chou, Katherine
and Hassidim, Avinatan
and Gokturk, Burak
and Vahdat, Amin
and Guan, Yuan
and Dhillon, Vikram
and Vaishnav, Eeshit Dhaval
and Lee, Byron
and Costa, Tiago R. D.
and Penad{\'e}s, Jos{\'e} R.
and Peltz, Gary
and Matias, Yossi
and Manyika, James
and Hassabis, Demis
and Xu, Yunhan
and Kohli, Pushmeet
and Pawlosky, Annalisa
and Karthikesalingam, Alan
and Natarajan, Vivek},
title={Accelerating scientific discovery with Co-Scientist},
journal={Nature},
year={2026},
month={May},
day={19},
issn={1476-4687},
doi={10.1038/s41586-026-10644-y},
url={https://doi.org/10.1038/s41586-026-10644-y}
}

@inproceedings{ICLR2025_ea94957d,
 author = {Si, Chenglei and Yang, Diyi and Hashimoto, Tatsunori},
 booktitle = {International Conference on Learning Representations},
 editor = {Y. Yue and A. Garg and N. Peng and F. Sha and R. Yu},
 pages = {94003--94092},
 title = {Can LLMs Generate Novel Research Ideas? A Large-Scale Human Study with 100+ NLP Researchers},
 url = {https://proceedings.iclr.cc/paper_files/paper/2025/file/ea94957d81b1c1caf87ef5319fa6b467-Paper-Conference.pdf},
 volume = {2025},
 year = {2025}
}

@misc{hou2026noveltyagentautonomousnoveltyreporting,
      title={NoveltyAgent: Autonomous Novelty Reporting Agent with Point-wise Novelty Analysis and Self-Validation}, 
      author={Jiajun Hou and Hexuan Deng and Wenxiang Jiao and Xuebo Liu and Xiaopeng Ke and Min Zhang},
      year={2026},
      eprint={2603.20884},
      archivePrefix={arXiv},
      primaryClass={cs.CL},
      url={https://arxiv.org/abs/2603.20884}, 
}

@misc{mostafa2026novelknowledgedrivenframeworkbiasaware,
      title={What Is Novel? A Knowledge-Driven Framework for Bias-Aware Literature Originality Evaluation}, 
      author={Abeer Mostafa and Thi Huyen Nguyen and Zahra Ahmadi},
      year={2026},
      eprint={2602.06054},
      archivePrefix={arXiv},
      primaryClass={cs.CL},
      url={https://arxiv.org/abs/2602.06054}, 
}

@misc{wu2026novbenchevaluatinglargelanguage,
      title={NovBench: Evaluating Large Language Models on Academic Paper Novelty Assessment}, 
      author={Wenqing Wu and Yi Zhao and Yuzhuo Wang and Siyou Li and Juexi Shao and Yunfei Long and Chengzhi Zhang},
      year={2026},
      eprint={2604.11543},
      archivePrefix={arXiv},
      primaryClass={cs.CL},
      url={https://arxiv.org/abs/2604.11543}, 
}

@article{scikit-learn,
    title={Scikit-learn: Machine Learning in {P}ython},
    author={Pedregosa, F. and Varoquaux, G. and Gramfort, A. and Michel, V.
            and Thirion, B. and Grisel, O. and Blondel, M. and Prettenhofer, P.
            and Weiss, R. and Dubourg, V. and Vanderplas, J. and Passos, A. and
            Cournapeau, D. and Brucher, M. and Perrot, M. and Duchesnay, E.},
    journal={Journal of Machine Learning Research},
    volume={12},
    pages={2825--2830},
    year={2011}
   }

@string{acl = {Association for Computational Linguistics}}

@string{anth = {https://aclanthology.org/}}

@inproceedings{su-etal-2025-many,title = "Many Heads Are Better Than One: Improved Scientific Idea Generation by A {LLM}-Based Multi-Agent System",author = "Su, Haoyang and Chen, Renqi and Tang, Shixiang and Yin, Zhenfei and Zheng, Xinzhe and Li, Jinzhe and Qi, Biqing and Wu, Qi and Li, Hui and Ouyang, Wanli and Torr, Philip and Zhou, Bowen and Dong, Nanqing",editor = "Che, Wanxiang and Nabende, Joyce and Shutova, Ekaterina and Pilehvar, Mohammad Taher",booktitle = "Proceedings of the 63rd Annual Meeting of the Association for Computational Linguistics (Volume 1: Long Papers)",month = jul,year = "2025",address = "Vienna, Austria",publisher = acl,url = anth # {2025.acl-long.1368/},doi = "10.18653/v1/2025.acl-long.1368",pages = "28201--28240",ISBN = "979-8-89176-251-0"}

@inproceedings{afzal-etal-2026-beyond,title = "Beyond ``Not Novel Enough'': Enriching Scholarly Critique with {LLM}-Assisted Feedback",author = "Afzal, Osama Mohammed and Nakov, Preslav and Hope, Tom and Gurevych, Iryna",editor = "Demberg, Vera and Inui, Kentaro and Marquez, Llu{\'i}s",booktitle = "Proceedings of the 19th Conference of the {E}uropean Chapter of the {A}ssociation for {C}omputational {L}inguistics (Volume 1: Long Papers)",month = mar,year = "2026",address = "Rabat, Morocco",publisher = acl,url = anth # {2026.eacl-long.121/},doi = "10.18653/v1/2026.eacl-long.121",pages = "2648--2671",ISBN = "979-8-89176-380-7"}

@inproceedings{amplayo-etal-2019-evaluating,title = "Evaluating Research Novelty Detection: Counterfactual Approaches",author = "Amplayo, Reinald Kim and Hwang, Seung-won and Song, Min",editor = "Ustalov, Dmitry and Somasundaran, Swapna and Jansen, Peter and Glava{\v{s}}, Goran and Riedl, Martin and Surdeanu, Mihai and Vazirgiannis, Michalis",booktitle = "Proceedings of the Thirteenth Workshop on Graph-Based Methods for Natural Language Processing (TextGraphs-13)",month = nov,year = "2019",address = "Hong Kong",publisher = acl,url = anth # {D19-5315/},doi = "10.18653/v1/D19-5315",pages = "124--133"}

@inproceedings{mysore-etal-2022-multi,title = "Multi-Vector Models with Textual Guidance for Fine-Grained Scientific Document Similarity",author = "Mysore, Sheshera and Cohan, Arman and Hope, Tom",editor = "Carpuat, Marine and de Marneffe, Marie-Catherine and Meza Ruiz, Ivan Vladimir",booktitle = "Proceedings of the 2022 Conference of the North American Chapter of the Association for Computational Linguistics: Human Language Technologies",month = jul,year = "2022",address = "Seattle, United States",publisher = acl,url = anth # {2022.naacl-main.331/},doi = "10.18653/v1/2022.naacl-main.331",pages = "4453--4470"}

@inproceedings{baek-etal-2025-researchagent,title = "{R}esearch{A}gent: Iterative Research Idea Generation over Scientific Literature with Large Language Models",author = "Baek, Jinheon and Jauhar, Sujay Kumar and Cucerzan, Silviu and Hwang, Sung Ju",editor = "Chiruzzo, Luis and Ritter, Alan and Wang, Lu",booktitle = "Proceedings of the 2025 Conference of the Nations of the Americas Chapter of the Association for Computational Linguistics: Human Language Technologies (Volume 1: Long Papers)",month = apr,year = "2025",address = "Albuquerque, New Mexico",publisher = acl,url = anth # {2025.naacl-long.342/},doi = "10.18653/v1/2025.naacl-long.342",pages = "6709--6738",ISBN = "979-8-89176-189-6"}

@inproceedings{lin-etal-2025-evaluating,title = "Evaluating and Enhancing Large Language Models for Novelty Assessment in Scholarly Publications",author = "Lin, Ethan and Peng, Zhiyuan and Fang, Yi",editor = "Jansen, Peter and Dalvi Mishra, Bhavana and Trivedi, Harsh and Prasad Majumder, Bodhisattwa and Hope, Tom and Khot, Tushar and Downey, Doug and Horvitz, Eric",booktitle = "Proceedings of the 1st Workshop on AI and Scientific Discovery: Directions and Opportunities",month = may,year = "2025",address = "Albuquerque, New Mexico, USA",publisher = acl,url = anth # {2025.aisd-main.5/},doi = "10.18653/v1/2025.aisd-main.5",pages = "46--57",ISBN = "979-8-89176-224-4"}

@inproceedings{li-etal-2025-chain-ideas,title = "Chain of Ideas: Revolutionizing Research Via Novel Idea Development with {LLM} Agents",author = "Li, Long and Xu, Weiwen and Guo, Jiayan and Zhao, Ruochen and Li, Xingxuan and Yuan, Yuqian and Zhang, Boqiang and Jiang, Yuming and Xin, Yifei and Dang, Ronghao and Rong, Yu and Zhao, Deli and Feng, Tian and Bing, Lidong",editor = "Christodoulopoulos, Christos and Chakraborty, Tanmoy and Rose, Carolyn and Peng, Violet",booktitle = "Findings of the Association for Computational Linguistics: EMNLP 2025",month = nov,year = "2025",address = "Suzhou, China",publisher = acl,url = anth # {2025.findings-emnlp.477/},doi = "10.18653/v1/2025.findings-emnlp.477",pages = "8971--9004",ISBN = "979-8-89176-335-7"}

@inproceedings{maiya-etal-2025-improving,title = "Improving Preference Extraction In {LLM}s By Identifying Latent Knowledge Through Classifying Probes",author = "Maiya, Sharan and Liu, Yinhong and Debnath, Ramit and Korhonen, Anna",editor = "Che, Wanxiang and Nabende, Joyce and Shutova, Ekaterina and Pilehvar, Mohammad Taher",booktitle = "Proceedings of the 63rd Annual Meeting of the Association for Computational Linguistics (Volume 1: Long Papers)",month = jul,year = "2025",address = "Vienna, Austria",publisher = acl,url = anth # {2025.acl-long.444/},doi = "10.18653/v1/2025.acl-long.444",pages = "9061--9081",ISBN = "979-8-89176-251-0"}

@inproceedings{yang-etal-2024-large-language,title = "Large Language Models for Automated Open-domain Scientific Hypotheses Discovery",author = "Yang, Zonglin and Du, Xinya and Li, Junxian and Zheng, Jie and Poria, Soujanya and Cambria, Erik",editor = "Ku, Lun-Wei and Martins, Andre and Srikumar, Vivek",booktitle = "Findings of the Association for Computational Linguistics: ACL 2024",month = aug,year = "2024",address = "Bangkok, Thailand",publisher = acl,url = anth # {2024.findings-acl.804/},doi = "10.18653/v1/2024.findings-acl.804",pages = "13545--13565"}

@inproceedings{maas-etal-2011-learning,title = "Learning Word Vectors for Sentiment Analysis",author = "Maas, Andrew L. and Daly, Raymond E. and Pham, Peter T. and Huang, Dan and Ng, Andrew Y. and Potts, Christopher",editor = "Lin, Dekang and Matsumoto, Yuji and Mihalcea, Rada",booktitle = "Proceedings of the 49th Annual Meeting of the Association for Computational Linguistics: Human Language Technologies",month = jun,year = "2011",address = "Portland, Oregon, USA",publisher = acl,url = anth # {P11-1015/},pages = "142--150"}

@inproceedings{he-etal-2024-decoding,title = "Decoding Probing: Revealing Internal Linguistic Structures in Neural Language Models Using Minimal Pairs",author = "He, Linyang and Chen, Peili and Nie, Ercong and Li, Yuanning and Brennan, Jonathan R.",editor = "Calzolari, Nicoletta and Kan, Min-Yen and Hoste, Veronique and Lenci, Alessandro and Sakti, Sakriani and Xue, Nianwen",booktitle = "Proceedings of the 2024 Joint International Conference on Computational Linguistics, Language Resources and Evaluation (LREC-COLING 2024)",month = may,year = "2024",address = "Torino, Italia",publisher = "ELRA and ICCL",url = anth # {2024.lrec-main.402/},pages = "4488--4497"}

@inproceedings{ju-etal-2024-large,title = "How Large Language Models Encode Context Knowledge? A Layer-Wise Probing Study",author = "Ju, Tianjie and Sun, Weiwei and Du, Wei and Yuan, Xinwei and Ren, Zhaochun and Liu, Gongshen",editor = "Calzolari, Nicoletta and Kan, Min-Yen and Hoste, Veronique and Lenci, Alessandro and Sakti, Sakriani and Xue, Nianwen",booktitle = "Proceedings of the 2024 Joint International Conference on Computational Linguistics, Language Resources and Evaluation (LREC-COLING 2024)",month = may,year = "2024",address = "Torino, Italia",publisher = "ELRA and ICCL",url = anth # {2024.lrec-main.722/},pages = "8235--8246"}

\appendix

\section{About Probing \acp{llm} During Generation}\label{sec:probing-rel-work}
Probing approaches quantify the extent to which \ac{llm} representations encode specific knowledge. While extensive research investigates internal knowledge across diverse domains such as sentiment \cite{maas-etal-2011-learning} and factual knowledge \cite{marks2024the}, spatial and temporal understanding \cite{gurnee2024language}, and world models \cite{li2023emergent}, such existing studies predominantly focus on layer-wise localization of internal knowledge \cite[\textit{inter alia}]{he-etal-2024-decoding,NEURIPS2024_b1b16c4b,ju-etal-2024-large,jin-etal-2025-exploring}. Work on probing \acp{llm} during different generation steps is scarce and primarily addresses steering text generation \cite[\textit{inter alia}]{10.1162/TACL.a.48,zur2025languagemodelsawareroad,klerings-etal-2025-steering}. Although some works investigate the encoded information in \acp{llm} before generating the first token \cite{gottesman-geva-2024-estimating,afzal-etal-2025-knowing}, they do not distinguish between functional phases during generation. We address this gap by providing the first comparison of \ac{llm} representations during the \textit{reasoning} (``thinking'') phase versus the \textit{response generation} phase via probing, demonstrating that \acp{llm} encode more information about research idea novelty judgments while \textit{thinking} than when producing the actual response. 

\begin{table}[h!]
\small
\centering
\renewcommand{\arraystretch}{1.1} 
\begin{tabularx}{\columnwidth}{|c|X|}
\hline
\multicolumn{1}{|c|}{\textbf{Score}} & \multicolumn{1}{c|}{\textbf{Degree of Novelty}} \\
\hline
\multirow{1}{*}{1} & The idea is not novel. All aspects already exist in prior work. \\ \hline
\multirow{2}{*}{2} & The idea is marginally novel. It represents only a minor variation of existing work. \\ \hline
\multirow{3}{*}{3} & The idea is somewhat novel. Aspects already exist in prior work. However, it might combine known approaches in new ways, apply them to new contexts, or propose incremental updates. \\ \hline
\multirow{2}{*}{4} & The idea is novel. It introduces new aspects not present in existing work. \\ \hline
\multirow{3}{*}{5} & The idea is highly innovative and novel. It is not present in existing work and potentially encourages new thinking or opens up new research directions. \\ \hline
\end{tabularx}
\caption{Novelty Judgment Rubric}
\label{tab:novelty-rubric}
\end{table}

\section{Evaluation Metrics}\label{sec:eval-metrics}

We briefly summarize the \acs{rino} evaluation metrics, which we adopt in this work. For full details, see \citet{schopf-etal-2026-is}. The metrics evaluate both numerical novelty scores and textual justifications.

\subsection{Novelty Score Metrics} 

We evaluate predicted novelty scores using macro-$F_1$, class-wise $F_1$, and \ac{mae}. Macro-$F_1$ measures overall classification performance across the five novelty categories, class-wise $F_1$ shows performance for each individual score, and \ac{mae} measures the average distance between predicted and human gold scores on the ordinal 1--5 scale.

\subsection{Justification Metrics} 

For textual justifications, \acs{rino} distinguishes between \textit{known aspects}, which describe overlaps with prior work, and \textit{novelty aspects}, which describe new contributions of the research idea. Following \acs{rino}, these metrics are computed using an \ac{llm}-as-a-judge approach that compares model-generated justifications against human gold-standard justifications. In this work, we use the GPT-OSS-120B \cite{openai2025gptoss120bgptoss20bmodel} model for evaluation.

\noindent\textbf{Alignment} Alignment measures whether the model-generated justification follows reasoning consistent with the human gold justification and supports a similar novelty judgment. Scores range from 0 to 1, where higher is better: 1 indicates strong agreement with the human rationale, while 0 indicates no alignment.

\noindent\textbf{Recall} Recall measures how many known-aspect and novelty-aspect arguments from the human gold justification are captured by the model-generated justification. Scores range from 0 to 100, where higher is better: 100 means all relevant gold arguments are covered, while 0 means none are covered.

\noindent\textbf{Additional Ratio} Additional ratio measures how many extra known-aspect and novelty-aspect arguments the model adds beyond the gold justification, while still being grounded in the related works or research idea. Scores are non-negative percentages, where 0\% means no additional grounded arguments are added and higher values indicate more extra grounded content. This metric is not inherently good or bad: moderate or high values can indicate useful additional evidence, but very high values may also reflect overly verbose justifications.

\noindent\textbf{Hallucination Rate} Hallucination rate measures the proportion of generated known-aspect and novelty-aspect arguments that are not supported by the related works or research idea. Scores range from 0\% to 100\%, where lower is better: 0\% indicates justifications that are fully grounded in the research idea and related works, while higher values indicate more unsupported or hallucinated content.

\section{LoRA Fine-tuning Details}\label{sec:fine-tune}

For the FineTune approach in Section \ref{sec:tpr}, we fine-tune the base \ac{llm} using Low-Rank Adaptation (LoRA; \citealp{hu2022lora}). We apply LoRA to all major projection layers in the transformer, including the query, key, value, and output projections of the attention mechanism, as well as the gate, up, and down projections in the feed-forward network. We use a rank of $r=16$, a scaling factor $\alpha=32$, and a LoRA dropout of $0.1$.
Training is performed for two epochs using a per-device batch size of 1 and gradient accumulation over 8 steps, resulting in an effective batch size of 8. We employ a learning rate of $2 \times 10^{-4}$ with a short warmup of 20 steps. To reduce memory consumption, gradient checkpointing is enabled, and training is conducted in bfloat16 precision.

\section{Experimental Details}
\label{app:experimental-details}
All experiments were conducted on two NVIDIA A100 (80GB) GPUs. The probing classifier was implemented using scikit-learn \cite{scikit-learn}. Hidden states were extracted using the Hugging Face Transformers library \cite{wolf-etal-2020-transformers}.

\section{On the Choice of Novelty Score Metrics}\label{sec:metrics-choice}

We use macro-$F_1$ as the primary metric for evaluating novelty score predictions, rather than \ac{mae}. Although \ac{mae} is useful for measuring the average ordinal distance between predicted and gold scores, it is less informative in our setting because the novelty scale is small (1--5) and model predictions are strongly concentrated around the middle categories. As shown in Table~\ref{tab:mae-scores}, different models and approaches obtain very similar \ac{mae} values, typically around one. This makes \ac{mae} unsuitable for evaluation in our setting.

\begin{table}[h!]
\centering
\footnotesize
\resizebox{\columnwidth}{!}{
\begin{tabular}{llccccccccc}
\toprule
& Model & Zero-shot & Few-shot & CoT & Moose & \begin{tabular}{@{}c@{}}Research \\ Agent\end{tabular} & \begin{tabular}{@{}c@{}}AI \\ Scientist\end{tabular} & \begin{tabular}{@{}c@{}}AI \\ Researcher\end{tabular} & FineTune & \ac{tpr} \\
\midrule
\multirow{5}{*}{\rotatebox{90}{Non-Reas.}}
& Gemma-3-4B & 1.0 & 0.9 & 1.0 & 0.9 & 0.9 & 0.9& 0.9 & 1.1 & 1.0 \\
& Gemma-3-12B & 1.0 & 0.9 & 0.9 & 1.0 & 0.9 & 0.9 & 0.9 & 1.1 & 1.1 \\
& Gemma-3-27B & 1.0 & 1.0 & 0.9 & 1.0 & 0.9 & 1.0 & 0.9 & 1.1 & 1.1 \\
& Llama-3.1-8B & 1.0 & 1.0 & 1.0 & 0.9 & 0.9 & 0.9 & 0.9 & 1.1 & 1.1 \\
& Llama-3.1-70B & 1.0 & 1.0 & 1.1 & 1.0 & 1.0 & 1.0 & 0.9 & 1.1 & 1.1 \\
\midrule
\multirow{4}{*}{\rotatebox{90}{Reas.}} 
& Qwen3-4B & 1.0 & 1.0 & 1.0 & 1.0 & 0.9 & 1.0 & 1.0 & 1.0 & 1.1 \\
& Qwen3-14B & 1.0 & 1.0 & 1.1 & 1.0 & 1.0 & 1.0 & 1.0 & 1.1 & 1.0 \\
& Qwen3-32B & 1.0 & 1.0 & 1.0 & 1.0 & 0.9 & 1.0 & 1.0 & 1.0 & 1.1  \\
& GPT-OSS-20B & 0.9 & 0.9 & 1.0 & 0.9 & 0.9 & 1.0 & 0.9 & 1.2 & 1.1  \\
\bottomrule
\end{tabular}
}
\caption{\ac{mae} scores for different approaches and \acp{llm} on the \acs{rino} test set.}
\label{tab:mae-scores}
\end{table}

The limitation arises because a model that repeatedly predicts a middle score, such as 3, can achieve a deceptively low \ac{mae}: many gold labels are only one or two points away on a five-point Likert scale. Conversely, a less biased model that also predicts more extreme novelty categories may occasionally incur larger absolute errors, even if it produces more accurate novelty judgments overall. Optimizing for \ac{mae} can therefore favor conservative middle-ground predictions, precisely the behavior we aim to mitigate.

Macro-$F_1$ better reflects our evaluation objective. It treats all novelty classes equally, regardless of their frequency, and explicitly rewards models for correctly predicting low, medium, and high novelty judgments. This is crucial for evaluating whether a model can correctly judge extreme novelty categories, such as \textit{``not novel''} and \textit{``highly novel''}, rather than merely staying close to the center of the scale. We therefore report \ac{mae} for completeness in Table~\ref{tab:mae-scores}, but use macro-$F_1$ and class-wise $F_1$ as the main indicators of novelty judgment performance.

\section{Justification Evaluation}\label{sec:justification-evaluation}
Beyond novelty judgment performance, we also evaluate the quality of textual justifications generated by open-source \acp{llm} using \ac{tpr}, with the results shown in Table \ref{tab:justification_metrics_tpr}. 

\begin{table}[ht!]
\centering
\footnotesize
\renewcommand{\arraystretch}{0.85}
\resizebox{\columnwidth}{!}{
\begin{tabular}{llccccccc}
\toprule
 & & ALI & \multicolumn{2}{c}{Recall} & \multicolumn{2}{c}{Add. Ratio} & \multicolumn{2}{c}{Hall. Rate} \\
\cmidrule(lr){4-5} \cmidrule(lr){6-7} \cmidrule(lr){8-9} 
& Model & & KA & NA & KA & NA & KA & NA \\
\midrule
\multirow{5}{*}{\rotatebox{90}{Non-Reas.}}
& Gemma-3-4B & 0.25 & 39.4 & 35.7 & 40.8 & 51.9 & 38.8 & 44.2 \\
& Gemma-3-12B & 0.39 & 58.2 & 50.4 & 56.7 & 72.4 & 8.6 & 6.8 \\
& Gemma-3-27B & 0.42 & 62.4 & 57.2 & 60.9 & 80.5 & 6.5 & 2.0 \\
& Llama-3.1-8B & 0.36 & 52.0 & 45.3 & 40.0 & 74.1 & 12.6 & 10.5 \\
& Llama-3.1-70B & 0.30 & 51.0 & 50.0 & 35.1 & 55.6 & 12.0 & 11.1 \\
\midrule
\multirow{4}{*}{\rotatebox{90}{Reas.}} 
& Qwen3-4B & 0.43 & 61.9 & 61.5 & 78.9 & 108.3 & 7.8 & 2.6 \\
& Qwen3-14B & 0.43 & 63.6 & \textbf{68.8} & \textbf{105.4} & \textbf{137.2} & 9.2 & 5.4 \\
& Qwen3-32B & \textbf{0.54} & \textbf{67.9} & 59.0 & 94.5 & 98.7 & \textbf{8.0} & \textbf{1.1} \\
& GPT-OSS-20B & 0.52 & 66.6 & 63.6 & 88.4 & 94.8 & 10.6 & 2.8 \\
\bottomrule
\end{tabular}
}\caption{Evaluation of textual justifications generated by different \acp{llm} using \ac{tpr}.}
\label{tab:justification_metrics_tpr}
\end{table}

\noindent\textbf{Alignment} The alignment scores of open-source models using \ac{tpr} are comparable to those of substantially larger proprietary \acp{llm} in Table~\ref{tab:benchmarking_results}. In particular, the strongest reasoning-capable models achieve ALI values around 0.5, close to the range observed for Claude and GPT models under zero-shot prompting. This indicates that \ac{tpr} does not merely improve numerical novelty prediction, but also enables open-source models to generate justifications whose reasoning remains broadly aligned with human gold-standard rationales.

\noindent\textbf{Recall} Similarly to propriety \acp{llm}, smaller open-source models exhibit relatively high recall, indicating substantial overlap between model-generated and human-annotated justification arguments. Comparing the results to the ones in Table~\ref{tab:benchmarking_results}, recall in open-source models is slightly lower than in the top-performing OpenAI models but remains competitive with Gemini~Pro models. 

\noindent\textbf{Additional Ratio} The \textit{Additional Ratio} is generally higher for reasoning-capable models than for non-reasoning models, indicating that reasoning models produce more elaborate justifications. While proprietary OpenAI \acp{llm} exhibit even higher additional ratios, the best-performing Qwen3 models remain competitive with Gemini Pro models, highlighting that \ac{tpr} enables smaller open-source models to generate rich novelty judgment justifications. 

\noindent\textbf{Hallucination Rate} Reasoning models show low hallucination rates similar to proprietary models, whereas non-reasoning models, particularly Gemma-3-4B, display higher hallucination rates when attempting to justify novelty judgments. This suggests that \ac{tpr} is most effective in producing grounded justifications when paired with reasoning \acp{llm}.

\noindent\textbf{Takeaway} Overall, \ac{tpr} allows open-source LLMs to generate high-quality, human-aligned novelty justifications, similar to what is achievable with large proprietary models. However, performance depends strongly on the model: \textit{reasoning-capable \acp{llm} consistently produce more accurate, elaborate, and reliable justifications, whereas smaller non-reasoning \acp{llm} may struggle to generate appropriate novelty judgment justifications.}

\begin{table*}[t]
\centering
\small
\renewcommand{\arraystretch}{1.2} 
\begin{tabularx}{\textwidth}{
    >{\centering\arraybackslash}m{1.1cm}   
    >{\centering\arraybackslash}m{1.1cm}   
    >{\raggedright\arraybackslash}m{6.0cm} 
    >{\raggedright\arraybackslash}m{6.0cm} 
}
\toprule
\textbf{Gold Novelty} & \textbf{Predicted Novelty} & \textbf{Gold Justification} & \textbf{LLM-generated Justification} \\
\midrule
$1$ & $3$ $\uparrow$ & The idea of \textcolor{lightgreen}{using neural networks to learn Green's functions is already known} and the proposed contribution is incremental, offering \textcolor{lightred}{no new aspects} beyond existing approaches. & The proposed idea combines fundamental solutions [...] \textcolor{lightred}{represents a somewhat novel synthesis}, but \textcolor{lightgreen}{the core components [...] are well-established in the literature.} \\
\hline
$2$ & $4$ $\uparrow$ & The approach adds [...] \textcolor{lightgreen}{extensions of existing communication mechanisms} rather than fundamentally new concepts, resulting in only \textcolor{lightred}{marginal novelty}. & The research idea proposes [...] \textcolor{lightred}{meaningful novelty} [...], though \textcolor{lightgreen}{it builds on existing concepts in communication} [...] \\
\hline
$3$ & $3$ $\checkmark$ & The approach primarily \textcolor{lightgreen}{assembles existing components} [...], resulting in a \textcolor{lightgreen}{somewhat novel} contribution. & The research idea \textcolor{lightgreen}{combines several existing concepts} [...] adds \textcolor{lightgreen}{incremental novelty}, [...] \\ 
\hline
$4$ & $4$ $\checkmark$ & The approach is \textcolor{lightgreen}{novel} [...] extending diffusion models beyond the usual focus on learning only the reverse process. [...] \textcolor{lightgreen}{has not been presented in prior work} [...] & [...] introduces a \textcolor{lightgreen}{novel approach} by jointly parameterizing both forward and reverse diffusion processes [...] \textcolor{lightgreen}{related works [...] primarily focus on the reverse process} [...] \\
\hline
$5$ & $4$ $\downarrow$ & The idea is \textcolor{lightred}{highly novel} because it uncovers a \textcolor{lightgreen}{previously unreported generalization phenomenon and establishes a new theoretical link} [...] & [...] \textcolor{lightgreen}{calibration literature doesn't explicitly connect ensemble disagreement to generalization} [...] \textcolor{lightred}{builds incrementally on existing calibration and ensemble concepts.} \\
\bottomrule
\end{tabularx}
\caption{Selected comparison of gold novelty judgments and LLM(Claude Opus 4.5)-generated novelty judgments. \textcolor{lightgreen}{Green} denotes alignment of model and gold justifications. \textcolor{lightred}{Red} highlights miscalibration, where the model's novelty judgment diverges from the gold judgment despite exhibiting a rationale aligned with the gold justification. Symbols indicate novelty overestimation ($\uparrow$), correct prediction ($\checkmark$), and underestimation ($\downarrow$).}
\label{tab:novelty_comparison}
\end{table*}

\begin{figure*}[h!]
\centering
\begin{mycode}[]{Zero-shot Research Idea Novelty Judgment Prompts}
system prompt = "You are an expert researcher experienced in judging the novelty of a research idea."

user prompt = f"""
    You are an expert in machine learning research evaluation. You will be given two inputs:

    1. A research idea with objective, problem statement, and solution approach.
    2. A list of related works, each with a title and abstract.

    Your task is to **assess the novelty of the research idea** compared to the related works.

    ### Instructions:
    - Analyze the research idea and summarize its key contributions.
    - Compare it with the related works to identify overlaps and differences.
    - Specifically, assess whether the idea introduces **significant new aspects** not present in existing work, or if it is largely a variation on known approaches.
    - Provide your output as a **JSON object only**, with:
      - "reasoning": a short paragraph (2-4 sentences) explaining the reasoning behind the novelty score.
      - "novelty_score": an integer between 1-5 where: {novelty_rubric}
      
    ### Inputs:

    **Research Idea:**
    {research_idea}

    **Related Works:**
    {related_works}

    ### Output Format:
    ```json
    {{
      "reasoning": <short explanation>,
      "novelty_score": <1|2|3|4|5>
    }}
    """
\end{mycode}
\caption{Prompts for the zero-shot approach to judging the novelty of research ideas. Here, an \ac{llm} receives a research idea, its related works, and the \acs{rino} novelty rubric, and is asked to generate both a numerical novelty score and a textual justification.}
\label{fig:zs-prompt}
\end{figure*}

\begin{figure*}[h!]
\centering
\begin{mycode}[]{TPR Instruction}
system prompt = "You are ReviewerGPT, an intelligent assistant that helps researchers evaluate the novelty of their ideas."

user prompt = f"""
    You are given some papers similar to the proposed idea (<IDEA> and </IDEA>). Your task is to evaluate the idea's novelty using the related papers (<PAPER> and </PAPER>) only.

    ### Novelty types:
    {novelty_class_descriptions}

    ### Instructions:
    - Use the example review below to write a review for the provided idea by comparing it to the related papers.
    - Don't assume any prior knowledge about the idea.
    - Make sure the generated review follows the format in example review provided below.
    - The review should be concise - around 60 to 100 words.

    ### Research Idea:
    {research_idea}

    ### Related Papers:
    {related_papers}

    ### Example Review:
    {example_review}

    ### Output Format:
    <REVIEW> concise review </REVIEW>

    Think step by step before generating the review!
    """
\end{mycode}
\caption{Instruction used for our \ac{tpr} approach as introduced in Section \ref{sec:tpr}. The instruction to reason step by step is included only for models that do not generate think tokens by default and is omitted otherwise.} 
\label{fig:tpr-prompt}
\end{figure*}

\end{document}